\documentclass[10pt,twocolumn,letterpaper]{article}

\usepackage[pagenumbers]{cvpr} 

\usepackage{graphicx}
\usepackage{amsmath}
\usepackage{epsfig}
\usepackage{amssymb}
\usepackage{times}
\usepackage{array}
\usepackage{booktabs}

\usepackage[pagebackref,breaklinks,colorlinks]{hyperref}
\usepackage[accsupp]{axessibility}  
\usepackage[sort, numbers]{natbib}
\usepackage{graphicx}
\usepackage{mathtools}
\usepackage{tabularx}
\usepackage{multirow}
\usepackage{dsfont}
\usepackage{enumitem}
\usepackage{bbm}
\usepackage{xcolor, colortbl}
\usepackage{wrapfig}
\usepackage{anyfontsize}
\usepackage[11pt]{moresize}

\usepackage[capitalize]{cleveref}
\crefname{section}{Sec.}{Secs.}
\Crefname{section}{Section}{Sections}
\Crefname{table}{Table}{Tables}
\crefname{table}{Tab.}{Tabs.}

\usepackage{enumitem}

\def\cvprPaperID{4410} %
\def\confName{CVPR}
\def\confYear{2022}

\usepackage{overpic}
\usepackage{enumitem} %
\usepackage{overpic} %
\usepackage{color}

\definecolor{turquoise}{cmyk}{0.65,0,0.1,0.3}
\definecolor{purple}{rgb}{0.65,0,0.65}
\definecolor{dark_green}{rgb}{0, 0.5, 0}
\definecolor{orange}{rgb}{0.8, 0.6, 0.2}
\definecolor{red}{rgb}{0.8, 0.2, 0.2}
\definecolor{darkred}{rgb}{0.6, 0.1, 0.05}
\definecolor{blueish}{rgb}{0.0, 0.3, .6}
\definecolor{light_gray}{rgb}{0.7, 0.7, .7}
\definecolor{pink}{rgb}{1, 0, 1}
\definecolor{greyblue}{rgb}{0.25, 0.25, 1}

\usepackage{blindtext}

\renewcommand{\paragraph}[1]{\vspace{1em}\noindent\textbf{#1}.}

\begin{document}

\definecolor{grey}{rgb}{0.9, 0.9, 0.9}
\newcommand{\ccol}{\cellcolor{grey}}
\definecolor{lblue}{rgb}{0, 0.2, 0.8}
\definecolor{orange}{rgb}{1.0, 0.5, 0.0}
\newcommand{\suha}[1]{{\color{lblue}{#1}}}
\newcommand{\suhac}[1]{{\color{orange}{(#1)}}}
\newcommand{\dhc}[1]{{\color{red}{#1}}}
\DeclarePairedDelimiter{\nint}\lfloor\rceil

\title{Semi-supervised Semantic Segmentation with Error Localization Network}

\author{
Donghyeon Kwon$^1$ \qquad
Suha Kwak$^{1,2}$ \qquad\\
Dept. of CSE, POSTECH$^1$ \qquad Graduate School of AI, POSTECH$^2$\\
{\tt\small \url{http://cvlab.postech.ac.kr/research/ELN/}}
}
\maketitle

\begin{abstract}\label{sec:abstract}

This paper studies semi-supervised learning of semantic segmentation, which assumes that only a small portion of training images are labeled and the others remain unlabeled. The unlabeled images are usually assigned pseudo labels to be used in training, which however often causes the risk of performance degradation due to the confirmation bias towards errors on the pseudo labels. We present a novel method that resolves this chronic issue of pseudo labeling. At the heart of our method lies error localization network (ELN), an auxiliary module that takes an image and its segmentation prediction as input and identifies pixels whose pseudo labels are likely to be wrong. ELN enables semi-supervised learning to be robust against inaccurate pseudo labels by disregarding label noises during training and can be naturally integrated with self-training and contrastive learning. Moreover, we introduce a new learning strategy for ELN that simulates plausible and diverse segmentation errors during training of ELN to enhance its generalization. Our method is evaluated on PASCAL VOC 2012 and Cityscapes, where it outperforms all existing methods in every evaluation setting. 

\end{abstract}      
\section{Introduction}
\label{sec:intro}
Recent advances in semantic segmentation have been attributed to supervised learning of deep neural networks~\citep{Fcn,deeplab_v2,deeplab_v3,deconvnet,PSPNet} on large-scale datasets~\citep{Pascalvoc,kitty,cityscapes,Mscoco}.
However, collecting training data for semantic segmentation is labour-intensive and time-consuming due to the prohibitive cost of pixel-wise class labeling, which often leads to a dataset limited in terms of the number of annotated data and class diversity. 
To address this issue, label efficient learning, such as semi-supervised learning~\cite{S4gan,Universal_SSL_seg,Naive_student,GCTNet,ECNet,Three_stage_self_training_for_ssl,Semantic_Seg_with_Generative_models,lai2021cac,he2021re,alonso2021semi}, unsupervised learning~\cite{Van_Gansbeke_2021_ICCV,Cho_2021_CVPR}, weakly supervised learning~\cite{affinitynet,IRNet,Huang_2018_CVPR,sppnet,wang2018weakly,sun2020mining,dong_2020_conta,chen2020weakly}, and synthetic-to-real domain adaptation~\cite{hoffman2018cycada,tsai2018learning,zou2018unsupervised,tsai2019domain,vu2019advent,li2019bidirectional,NEURIPS2020_243be281}, has been proposed for semantic segmentation.

This paper studies semi-supervised learning of semantic segmentation, which assumes that only a subset of training images are assigned segmentation labels while the others remain unlabeled.
Undoubtedly, the key to the success of this task is to utilize the unlabeled images effectively.
Self-training~\cite{GCTNet,ECNet, Xie_2020_CVPR, Naive_student} and contrastive learning~\cite{lai2021cac, Pixel_contra_SSL, alonso2021semi, C3seg} are techniques commonly used for the purpose in literature. Self-training generates pseudo labels of unlabeled images using a model trained on labeled ones, and uses them for supervised learning. Meanwhile, contrastive learning forces feature vectors corresponding to the same pseudo label to be close to each other. Although these techniques have improved the performance of semi-supervised semantic segmentation substantially, they share a common drawback:
Since predictions for unlabeled images are usually corrupted by errors, learning using such predictions as supervision causes \emph{confirmation bias} towards the errors and returns corrupted models consequently. Most existing methods alleviate this issue simply by not using uncertain predictions as supervision~\cite{S4gan, lai2021cac, he2021re, alonso2021semi}, but their performance depends heavily on hand-tuned thresholds.

A recent approach deals with errors on pseudo labels by learning and exploiting an auxiliary network that corrects the errors~\cite{ECNet,GCTNet}; this model, called error correction network (ECN), learns from the difference between predictions of the main segmentation network and their ground truth labels on the labeled subset of training images. 
Ideally, ECN can significantly improve the quality of pseudo labels, but in practice, its advantage is often limited due to the challenges in its training. 
Since the segmentation network is quickly overfitted to a small number of labeled images, its outputs used as input to ECN do not cover a wide variety of prediction errors that ECN faces in testing, which results in limited generalization capability of ECN. 

\begin{figure*}[t]
    \centering
    \includegraphics[width=\linewidth]{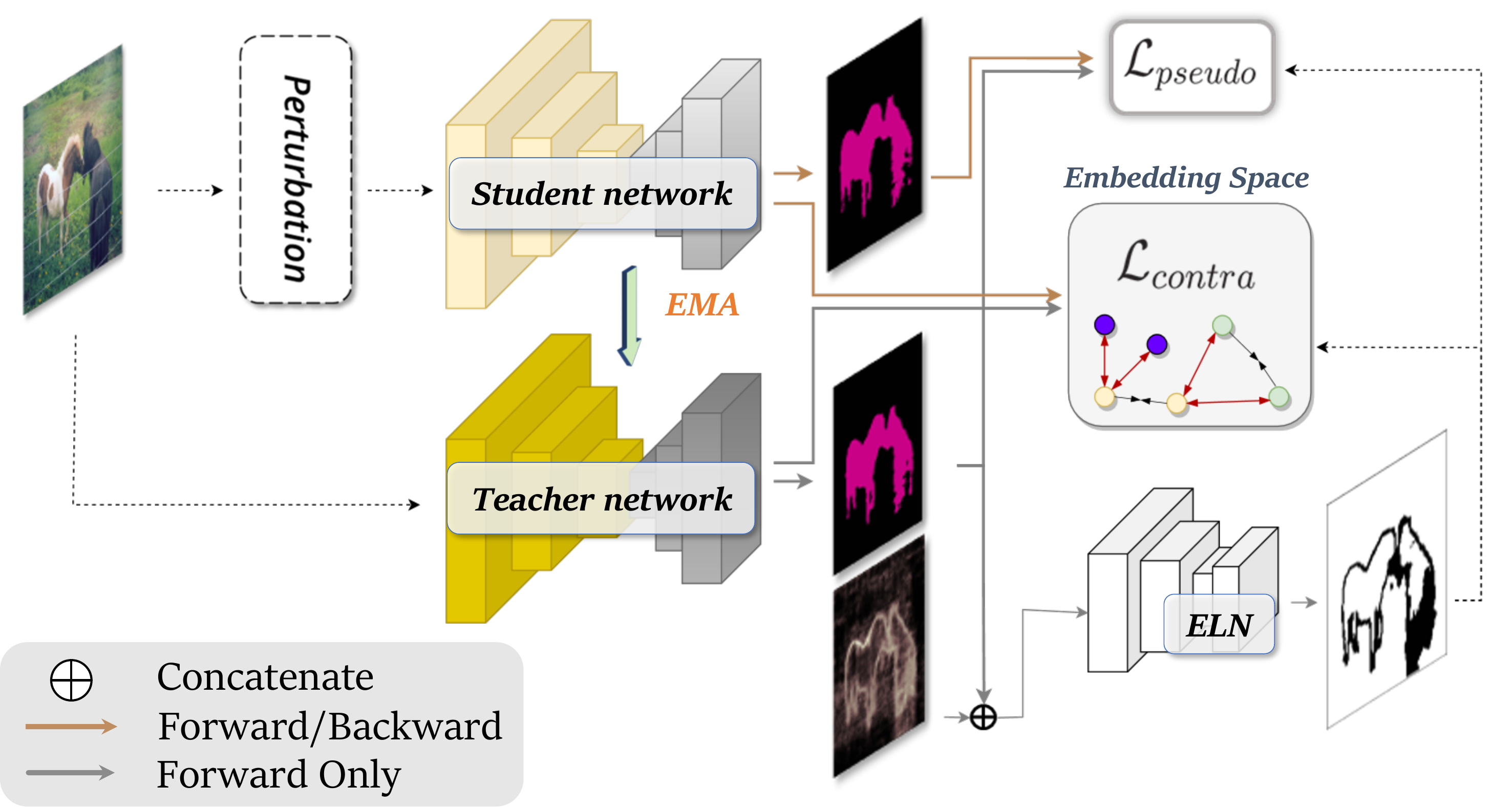}
    \caption{
    Our semi-supervised learning framework incorporating ELN. It employs two segmentation networks, the student ($s$), which will be our final model, and the teacher ($t$) used for generating pseudo labels. The student is trained using the pseudo labels of the teacher in two different ways, self-training and contrastive learning. 
    To be specific, the decoder has two heads, one for segmentation (\emph{Seg}) and the other for feature embedding (\emph{Proj}); self-training and contrastive learning are applied to outputs of the \emph{Seg} and \emph{Proj} heads, respectively.
    Then the teacher is updated by an exponential moving average (EMA) of the student.
    ELN allows both self-training and contrastive learning to be robust against noises on pseudo labels by identifying and disregarding pixels whose pseudo labels are likely to be noisy.
    } 
    \label{fig:semi}%
\end{figure*}

We present a novel method that is also dedicated to handling errors on pseudo labels yet better generalizes to those of arbitrary unlabeled images. 
The core of our method is the error localization network (ELN), which identifies pixels with erroneous pseudo labels in the form of binary segmentation.
As will be demonstrated empirically, simply disregarding invalid pseudo labels, instead of correcting them, is sufficient to alleviate the confirmation bias and to learn accurate segmentation models. 
More importantly, since error localization is a class-agnostic subproblem of error correction and accordingly easier to solve, 
it is more straightforward to train an accurate and well-generalizable network for the target task.

Moreover, we design a novel training strategy for ELN to further improve its generalization.
Specifically, we attach multiple auxiliary decoders to the main segmentation network and train them to achieve different accuracy levels so that they simulate the segmentation network at different training stages.
ELN is then trained to localize errors on the predictions given by the auxiliary decoders as well as the main segmentation network.
This strategy improves generalization of ELN since such predictions used as input to ELN potentially exhibit error patterns that the segmentation network causes during self-training with unlabeled images.

The trained ELN is then used for semi-supervised learning of semantic segmentation; the overall pipeline incorporating ELN is illustrated in Fig.~\ref{fig:semi}.
Our framework exploits unlabeled images in two ways: self-training and contrastive learning, both relying on pseudo labels.
To this end, we adopt two segmentation networks: A student network, which will be our final model, and a teacher network generating pseudo labels and updated by an exponential moving average of the student.
Self-training is done by learning the student using pseudo labels produced by the teacher.
Meanwhile, contrastive learning encourages embedding vectors of the student and teacher to be similar if their pseudo labels are identical. 
ELN helps improve the effect of both self-training and contrastive learning by filtering out potentially erroneous pseudo labels.

Following the convention, the proposed method is evaluated on the PASCAL VOC 2012~\cite{Pascalvoc} and Cityscapes~\cite{cityscapes} datasets while varying the number of labeled training images,
and it demonstrates superior performance to previous work on both of the datasets.

\begin{figure*}[t]
    \centering
    \includegraphics[width=\textwidth]{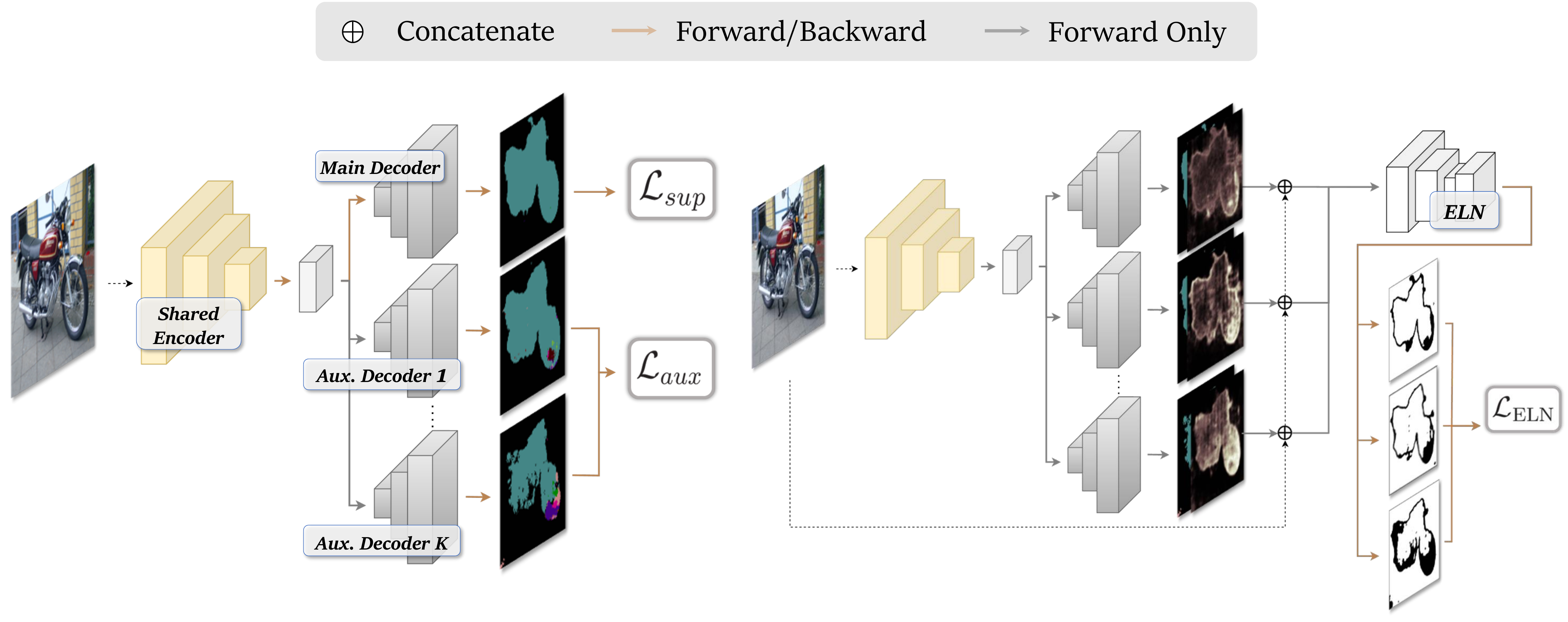}
    \caption{
    Training ELN along with the main segmentation network and the auxiliary decoders. (\emph{left}) The main segmentation network is trained with the ordinary cross-entropy loss $\mathcal{L}_{sup}$, but the auxiliary decoders are trained with constrained cross-entropy losses $\mathcal{L}_{aux}$ so that they are inferior to the main segmentation network, and their predictions contain plausible and diverse errors intentionally. 
    (\emph{right}) All predictions from the decoders are used as training input to ELN, which learns to localize errors on the predictions. 
    Note that ELN and other components are trained simultaneously, although their training processes are drawn separately in this figure for brevity.
    }
    \label{fig:sup}%
\end{figure*}

In brief, our main contribution is three-fold.
\vspace{-2mm}
\begin{itemize}[leftmargin=5mm] 
\itemsep=-0.5mm
    \item We propose \emph{error localization}, a new approach to dealing with errors on pseudo labels. It is simple yet effective and can be naturally incorporated with self-training and contrastive learning. Moreover, we empirically demonstrate the superiority of error localization to error correction.
    \item We develop a new strategy for generating diverse and plausible prediction errors intentionally during the training of ELN. This improves the generalization of ELN even using a small number of labeled data for training.
    \item Segmentation networks trained by our method achieves the state of the art on two benchmark datasets, PASCAL VOC 2012 and Cityscapes, in every setting.
\end{itemize}

\section{Related Work}
\label{sec:relatedwork}

\noindent \textbf{Semantic segmentation.}
The goal of semantic segmentation is to generate dense pixel-wise classification.
Starting with FCN \cite{hoffman2016fcns}, which replaced the classifier’s last fully-connected layer with a fully convolutional layer for the first time, various approaches have been studied early.
An encoder-decoder structure has been proposed to obtain an accurate high-resolution output \cite{deconvnet,unet}, and structures such as ASPP \cite{deeplab_v2} and PSPNet \cite{PSPNet} have been exploited to obtain more diverse spatial contexts.
An attention mechanism has been studied to obtain a global relation \cite{Ccnet,DualAttentionNetwork}.
However, the success of these models requires a large amount of data, which costs expensive labour.

\noindent \textbf{Semi-supervised semantic segmentation.} 
Attempts to reduce the cost by applying a semi-supervised learning scheme have been studied intensely. Several methods \cite{Universal_SSL_seg, Semantic_Seg_with_Generative_models,S4gan} based on GAN and adversarial learning have been studied to reduce the gap between prediction on unlabeled and labeled data.
One of the techniques frequently used in semi-supervised learning  \cite{Fixmatch, SSL_with_cross_consistency, SSL_need_strong_perturbation, Three_stage_self_training_for_ssl} is consistency regularization.
It allows the decision boundary to be located in a low-density region by using constraints to make the outputs of various perturbed inputs consistent with each other.
Another approach~\cite{ECNet, GCTNet, Xie_2020_CVPR, Naive_student}, self-training, is a method of generating pseudo labels with unlabeled data by pre-trained model and training the model with both labeled and pseudo labeled data.
Recently, various methods~\cite{lai2021cac, Pixel_contra_SSL, alonso2021semi, C3seg} have applied contrastive learning~\cite{contrastive_learning} to semantic segmentation in a semi-supervised manner, showing significant performance improvement. 

\noindent \textbf{Self-correction networks for semi-supervised semantic segmentation.} 
The idea of correcting pseudo labels by an auxiliary network has been studied in~\cite{GCTNet, ECNet}.
They presented networks that correct errors of pseudo labels by learning the difference between
predicted and ground truth segmentation labels.
However, it is challenging to train such networks effectively in the semi-supervised learning setting since the segmentation network is quickly overfitted to the labeled data, leading to a poor generalization of correction networks. 
To address this generalization issue of the previous work, we introduce a new auxiliary task called error localization, and present ELN and its training strategy.

\section{Proposed Method}
\label{sec:Method}

Our framework consists of two-stage, learning ELN using labeled data and semi-supervised learning with ELN.
The major issue in the first stage is the lack of diversity in predictions of the main network, which leads to a poor generalization of ELN.
To address this issue, in addition to the main segmentation network (encoder $\mathcal{E}$, decoder $\mathcal{D}$), we employ auxiliary decoders ($\mathcal{D}_{aux}^1$, ..., $\mathcal{D}_{aux}^{K}$) that are learned to be inferior to the main segmentation network intentionally; predictions of the auxiliary decoders will depict plausible and diverse errors. 
ELN is learned along with the segmentation network and auxiliary decoders to identify errors on their predictions. 
The overall procedure of ELN training is illustrated in Fig.~\ref{fig:sup}.

In the second stage, the trained ELN is then used for semi-supervised learning of semantic segmentation, where unlabeled images are exploited in two ways, self-training, and contrastive learning.
The role of ELN is to identify pixels whose pseudo labels are likely to be erroneous so that we disregard such pixels in the process of self-training and contrastive learning for stable and effective training. 

The remainder of this section presents details of the two stages of our method.

\subsection{Learning ELN Using Labeled Data} 
\label{sec:Supervised training stage}

At first, the main segmentation network is pre-trained with the standard pixel-wise cross-entropy loss $\mathcal{L}_{sup}$ on the set of labeled images $D_L$. Let $\mathcal{L}_{ce}(P, Y)$ denote the standard pixel-wise cross-entropy between segmentation prediction $P$ and its ground truth label $Y$:
\begin{equation}
    \mathcal{L}_{ce}(P, Y) = -\sum_{i}{Y_i^{\top}\log(P_i)},
\end{equation}
where $i$ is the index indicating each pixel of the input and $Y_i$ is the one-hot vector of the ground truth for pixel $i$.
Let $P=\mathcal{D}(\mathcal{E}(X))$ denote segmentation prediction of the main network for an image $X$. $\mathcal{L}_{sup}$ is then given by
\begin{equation}
    \mathcal{L}_{sup} = \frac{1}{| D_L |}\sum_{X \in D_L}\mathcal{L}_{ce}(P, Y),
\end{equation}
where $Y$ is the ground truth of the input image $X$.

When the pre-training is completed, each auxiliary decoder is trained similarly to the main network but with a constrained cross-entropy loss, which is minimized only up to a certain multiple of $\mathcal{L}_{ce}(P, Y)$ and its gradient is not propagated beyond the auxiliary decoder. 
Let $K$ be the number of all auxiliary decoders and $k$ be their index.
Then the total loss for the $K$ auxiliary decoders, denoted by $\mathcal{L}_{aux}$, is given by
\begin{equation}
    \begin{split}
        \mathcal{L}_{aux} =& \frac{1}{| D_L |}\sum_{X \in D_L}  
        \sum_{k=1}^{K} \\
        &\mathds{1}\{ \mathcal{L}_{ce}(P^k, Y) > 
        \alpha^k \cdot \mathcal{L}_{ce}(P, Y) \}\cdot \mathcal{L}_{ce}(P^k, Y),
    \end{split}
\end{equation}
where $P^k=\mathcal{D}^k(\mathcal{E}(X))$ denotes segmentation prediction of the $k$th decoder, and $\alpha^k$ indicates a scale hyper-parameter for constraining the loss applied to the $k$th auxiliary decoder. Training the auxiliary decoders in this way enables them to produce plausibly erroneous predictions, which are used as training input to ELN.

Given an image and its segmentation prediction as input, ELN is trained to localize errors on the prediction through supervised learning, where true locations of the errors are revealed by comparing the prediction with its ground truth counterpart.
Let $E^k$ be the pixel-wise entropy map of $P^k$ and $B^k = \text{ELN}(X \oplus P^k \oplus E^k)$ denote the prediction of ELN in the form of binary segmentation map, where $\oplus$ represents channel-wise matrix concatenation. Then the binary cross-entropy loss for ELN, $\mathcal{L}_{\text{ELN}}$, is given by
\begin{equation}
    \begin{split}
        \mathcal{L}_{\text{ELN}} = \frac{1}{|D_L|\cdot (K+1) }\sum_{X \in D_L}\sum_{k=0}^{K} \mathcal{L}_{ce}(B^k, M^k),
    \end{split}\label{eq:ELN}
\end{equation}
where $M^k$ denotes the ground truth mask of $B^k$; $M_i^k$ is 1 if the prediction of pixel $i$ is correct and 0 otherwise. Note that $k=0$ denotes the main decoder.

Despite using the auxiliary decoders, the population of pixel-level binary labels in $M^k$ is typically biased to 1 (correct), which impairs the error identification ability of ELN.
To alleviate this, a re-weighting factor is applied to pixels with incorrect predictions in $\mathcal{L}_{\text{ELN}}$ for balanced training.
Let $\mathcal{L}_{wce}$ denote a weight re-adjusted pixel-wise cross-entropy between segmentation prediction $P$ and its binary ground truth label $Y$:
\begin{equation}
    \begin{split}
    \mathcal{L}_{wce}(P, Y) = \phantom{AAAAAAAAAAAAA}\\
    -\sum_{i}\bigg(\mathds{1}\{Y_i = 0\} \frac{\sum_{j}\mathds{1}\{Y_j = 1\}}{\sum_{j}\mathds{1}\{Y_j = 0\}}&Y_i^{\top}\log(P_i) \\
    + \, \mathds{1}\{Y_i = 1\} &Y_i^{\top}\log(P_i) \bigg).
    \end{split}
\end{equation}
Then the loss in Eq.~(\hyperref[eq:ELN]{4}) is revised as  
\begin{equation}
    \begin{split}
        \mathcal{L}_{\text{ELN}} = \frac{1}{|D_L|\cdot (K+1) }\sum_{X \in D_L}\sum_{k=0}^{K} \mathcal{L}_{wce}(B^k, M^k). 
\end{split}
\end{equation}

The total loss minimized in the first stage for labeled data is as follows:
\begin{equation}
    \mathcal{L}_{labeled} = \mathcal{L}_{sup} + \mathcal{L}_{aux} + \mathcal{L}_{\text{ELN}}\,.
\end{equation}
Note that losses in $\mathcal{L}_{labeled}$ are jointly optimized in the first stage, although in pre-training $\mathcal{L}_{sup}$ is solely minimized.

\subsection{Semi-supervised Learning with ELN} 
After learning ELN, the main segmentation network is trained on the set of unlabeled images $D_U$ with two losses, a self-training loss and a pixel-wise contrastive loss. We adopt the mean teacher framework \cite{MeanTeacher}, 
which allows the teacher network to provide more stable pseudo supervision to the student network.
Weights $\Tilde{\theta}$ of the teacher ($\Tilde{\mathcal{E}}, \Tilde{\mathcal{D}}$) are updated by the exponential moving average of weights $\theta$ of the student ($\mathcal{E}, \mathcal{D}$) with an update ratio $\beta$: 
\begin{equation}
    \Tilde{\theta}_t = \beta{\Tilde{\theta}_{t-1}} + (1 - \beta)\theta_t \,.
\end{equation}

The proposed self-training loss $\mathcal{L}_{pseudo}$ is the pixel-wise cross-entropy loss like $\mathcal{L}_{sup}$, but is applied only to valid pixels identified by ELN. 
Let $\Tilde{P} = \Tilde{\mathcal{D}}(\Tilde{\mathcal{E}}(X))$ denote segmentation prediction of the teacher network and $P^a = \mathcal{D}(\mathcal{E}(A \cdot X))$ denote that of the student network, where $A$ is the perturbation operator applied to the input image $X$. Also, let $\Tilde{B} = \text{ELN}(X \oplus \Tilde{P} \oplus \Tilde{E})$ be the binary segmentation output of ELN.
Then $\mathcal{L}_{pseudo}$ is given by
\begin{equation}
    \begin{split}
        \mathcal{L}_{pseudo} = -\frac{1}{| D_U |}\sum_{X \in D_U} \sum_{i}\nint{\Tilde{B}}_i\cdot \hat{Y}_i^{\top}\log(P^a_i),
    \end{split}
\end{equation}
where $\nint{}$ is a function rounding to the nearest integer and $\hat{Y}_i$ denotes the one-hot vector of the pseudo label for pixel $i$.
Through the rounded binary mask, the main segmentation network can be trained on valid pixels only.

In order to further improve the quality of learned features, we adopt a pixel-wise contrastive loss $\mathcal{L}_{contra}$. Specifically, in this loss,
features whose pseudo labels are the same attract each other while those from different categories are pushed away in the feature space. Instead of applying the loss on a single image, we expand its range to the whole input batch for considering various feature relations, leading to a significant performance improvement.
For a given input, let $\Omega^i_p$ denote a set of pixels belonging to the class of pixel $i$ and $\Omega^i_n$ denote a set of pixels that do not belong to the class of pixel $i$. Also, let $\mathbf{d}$ represent a distance function, $\mathbf{d}(f_1, f_2) = \text{exp}(\text{cos}(f_1, f_2)/\tau)$, where $\text{cos}$ means the cosine similarity and $\tau$ is a temperature hyper-parameter. The pixel-wise contrastive loss $\mathcal{L}_{contra}$ is then given by
\begin{equation}
    \begin{split}
        \mathcal{L}_{contra} =&\\
        -\frac{1}{|V|}\sum_{i \in V}&\sum_{j \in \Omega^i_p}
        \log\frac{\mathbf{d}(f_i,\Tilde{f}_j)}{\mathbf{d}(f_i,\Tilde{f}_j) + \sum_{k \in \Omega^i_n} \mathbf{d}(f_i,\Tilde{f}_k)},
    \end{split}
\end{equation}
where $V$ denotes the set of valid pixels on $D_U$, $f_i$ and $\Tilde{f}_i$ are feature embeddings of pixel $i$ from the student and teacher networks, respectively.

The total loss for unlabeled data is as follows:
\begin{equation}
    \mathcal{L}_{unlabeled} = \mathcal{L}_{pseudo} + \mathcal{L}_{contra} \,.
    \label{eqn:11}
\end{equation}
Note that labeled data are also involved in training through $\mathcal{L}_{labeled}$.
When training is completed, only the student network is used at inference since the others, including ELN, are all auxiliary modules that support semi-supervised learning of the student.

\section{Experiments}
\label{sec:Experiments}
\subsection{Network Architecture} 
We use DeepLab v3+~\cite{deeplabv3plus2018} with ResNet~\cite{resnet} backbone as our segmentation network 
since it has been adopted in recent papers~\cite{ECNet, lai2021cac, alonso2021semi} 
and shares a similar structure with Deeplab v2~\cite{deeplab_v2}, that has been widely used in literature~\cite{SSL_need_strong_perturbation, S4gan, GCTNet, ClassMix}.

The proposed model mainly consists of two types of networks, the main segmentation network and the ELN. Each network is formed with an encoder and decoder. The encoder includes a ResNet~\cite{resnet} backbone, and the decoder (including the auxiliary decoders) contains sub-modules such as an atrous spatial pyramid pooling layer~\cite{deeplab_v2}, a pixel-wise classifier for segmentation (\emph{Seg} in Fig.~\ref{fig:semi}), and a projector for feature embedding (\emph{Proj} in Fig.~\ref{fig:semi}). The last two modules are implemented by two $1 \times 1$ convolutional layers and one intermediate ReLU activation layer.

We adopt ResNet-50 or ResNet-101 as the backbone of the main network, and ResNet-34 for ELN.
The backbones are pre-trained on ImageNet, but
since the input to ELN is the concatenation of an image and tensors, its first convolutional layer is accordingly re-designed and fine-tuned.

\begin{table}

\centering
\resizebox{\linewidth}{!}{ %
\begin{tabular}{@{}lcccccc@{}}
\toprule
Method & SegNet & Backbone & 1/20 & 1/8 & 1/4 & Full\\
\midrule
CutMix \cite{SSL_need_strong_perturbation} & DL2 & R101 & 66.48 &  67.60 & - & 72.54 \\ %
S4GAN+MLMT \cite{S4gan} & DL2 & R101 & 62.9 &  67.3 & - & 73.2\\ %
GCT \cite{GCTNet} & DL2 & R101 & - &  72.14 & 73.62 & 75.73\\ %
Alonso et al.~\cite{alonso2021semi} & DL2 & R101 & 67.8 & 69.9 & - & 72.6\\ %
\midrule
Baseline & DL3+ & R50 & 59.88 &  67.63 & 70.56 & 76.6\\ 
ECS~\cite{ECNet} & DL3+ & R50 & - &  70.22 & 72.60 & 76.29\\ %
Xin et al.~\cite{lai2021cac} & DL3+ & R50 & - &  72.4 & 74.0 & 76.5\\ %
Alonso et al.~\cite{alonso2021semi} & DL3+ & R50 & 69.1 & 71.8 & - & 75.9\\ %
\toprule
\textbf{Ours} & DL3+ & R50 & 70.52 & 73.20 & 74.63 & -\\ 
\midrule
\midrule
Baseline & DL3+ & R101 & 64.47  & 69.52 & 72.95 & 78.24\\ 
CutMix~\cite{SSL_need_strong_perturbation} & DL3+ & R101 & 69.57  & 72.45 & - & 76.73 \\ %
Xin et al.~\cite{lai2021cac} & DL3+ & R101 & -  & 74.6 & 76.3 & 78.2\\ %
\toprule
\textbf{Ours} & DL3+ & R101 & \textbf{72.52} & \textbf{75.10} & \textbf{76.58} & - \\
\bottomrule
\end{tabular}
} %
\caption{
mIoU value in the PASCAL VOC 2012 \emph{val} set with different labeled-unlabeled ratios. All results of our experiments are averaged from three different subsets of the same ratio.
} %
\label{sec:VOC-RESULT} 
\end{table}

\begin{table}
\vspace{5mm}
\centering
\resizebox{\linewidth}{!}{ %
\begin{tabular}{@{}lccccccc@{}}
\toprule
Method & SegNet & Backbone & 1/8 & 1/4 & 1/2 & Full\\
\midrule
CutMix~\cite{SSL_need_strong_perturbation} & DL2 & R101 & 60.34 & 63.87 & - & 67.68 \\ %
S4GAN~\cite{S4gan} & DL2 & R101 & 59.3 & 61.9 & - & 65.8\\ %
Alonso et al.~\cite{alonso2021semi} & DL2 & R101 & 63.0 & 64.8 & - & 66.4\\ %
\midrule
Baseline & DL3+ & R50 & 59.88 & 61.86 & 67.63 & 77.70\\ 
ECS~\cite{ECNet} & DL3+ & R50 & 67.38 & 70.70 & 72.89 & 74.76 \\ %
Xin et al.~\cite{lai2021cac} & DL3+ & R50 & 69.7 & 72.7 & - & 77.5 \\ %
Alonso et al.~\cite{alonso2021semi} & DL3+ & R50 & 70.0 & 71.6 & - & 74.2 \\ %
\toprule
\textbf{Ours} & DL3+ & R50 & \textbf{70.33}& \textbf{73.52} & \textbf{75.33} & -\\ 
\bottomrule
\end{tabular}
} %
\caption{
mIoU value in the Cityscapes \emph{val} set with different labeled-unlabeled ratios. All results of our experiments are averaged from three different subsets of the same ratio.
} 
\label{sec:CITY-RESULT}
\end{table}
\subsection{Implementation Details} 
\noindent\textbf{Datasets.} We conduct experiments on two different datasets, PASCAL VOC 2012~\cite{Pascalvoc} and Cityscapes~\cite{cityscapes}. PASCAL VOC 2012 is a standard semantic segmentation dataset consisting of 21 classes including the background class. The dataset has three separate subsets for training, validation, and testing; the subsets consist of 1464, 1449, 1456 images, respectively. Following the common practice, we use additional 9118 training images from the Segmentation Boundary (SBD) Dataset~\cite{Hariharan}. During training on PASCAL VOC 2012, we resize images to $512 \times 512$ pixels. 
Cityscapes~\cite{cityscapes} is a dataset of urban driving scenes with 19 classes for objects and background stuffs. 
It consists of training, validation, and testing splits with 2975, 500, and 1525 images, respectively. 
Images of the dataset is randomly cropped to $512 \times 1024$.

\noindent\textbf{Data augmentation.}\, Random horizontal flip is applied to both training datasets with the probability of 0.5. As the perturbation operator for the semi-supervised learning, we adopt color jittering and random grayscale with the probability of 0.2.

\noindent\textbf{Optimizer.}\, AdamW~\cite{adamw} is adopted with learning rate 1e-4 and weight decay 1e-5. 

\noindent\textbf{Hyper-parameters.}\, For both labeled and unlabeled data, the size of a mini-batch is 6 on PASCAL VOC 2012 and 4 on Cityscapes.
We assign 20 and 50 to the first and second auxiliary decoders, respectively. The temperature value $\tau$ of $\mathcal{L}_{contra}$ is set to 0.5. The update ratio $\beta$ is set to 0.995. 

\noindent\textbf{Evaluation Metrics.}\, We adopt the mean Intersection-over-Union (mIoU) as an evaluation metric.
During evaluation, image of PASCAL VOC 2012 are resized to $512 \times 512$ and those of Cityscapes are used as-is. 
We conduct experiments on several proportions of labeled data to unlabeled data for validating our method under different conditions. For PASCAL VOC 2012, we use three ratios, 1/20, 1/8, and 1/4, while 1/8, 1/4, and 1/2 are used for Cityscapes.

\subsection{Results}

\begin{figure*}[t!]
\centering
    \setlength{\tabcolsep}{0.05pt}
    \renewcommand{\arraystretch}{0.01}
    \resizebox{\textwidth}{!}{ %
        \begin{tabular}{ccccc}
              \includegraphics[width=3.0em, height=2.80em]{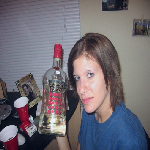} 
            & \includegraphics[width=3.0em, height=2.80em]{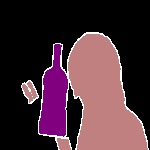}  
            & \includegraphics[width=3.0em, height=2.80em]{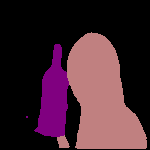} 
            & \includegraphics[width=3.0em, height=2.80em]{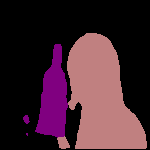} 
            & \includegraphics[width=3.0em, height=2.80em]{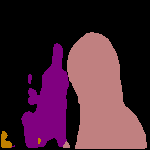} 
            \\
              \includegraphics[width=3.0em, height=2.80em]{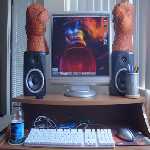} 
            & \includegraphics[width=3.0em, height=2.80em]{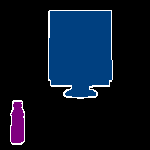}  
            & \includegraphics[width=3.0em, height=2.80em]{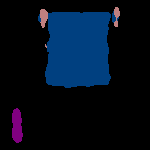} 
            & \includegraphics[width=3.0em, height=2.80em]{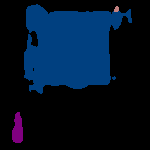} 
            & \includegraphics[width=3.0em, height=2.80em]{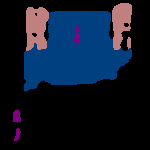} 
            \\
              \includegraphics[width=3.0em, height=2.80em]{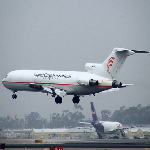} 
            & \includegraphics[width=3.0em, height=2.80em]{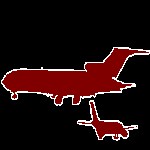}  
            & \includegraphics[width=3.0em, height=2.80em]{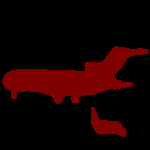} 
            & \includegraphics[width=3.0em, height=2.80em]{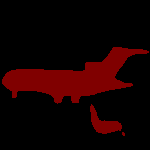} 
            & \includegraphics[width=3.0em, height=2.80em]{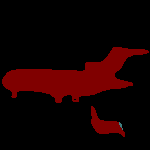} 
            \\
            &&&&\\
            &&&&\\
              {\fontsize{3}{4}\selectfont Input Image}
            & {\fontsize{3}{4}\selectfont GT}
            & {\fontsize{3}{4}\selectfont 1/4}
            & {\fontsize{3}{4}\selectfont 1/8}
            & {\fontsize{3}{4}\selectfont 1/20}
            \\

        \end{tabular}
    }
    \caption{Qualitative results on a \emph{val} set of PASCAL VOC 2012 in various proportions of labeled data to unlabeled data.}
    \label{sec:qual-2-voc}
\end{figure*}

\begin{figure*}[t!]
\centering
    \setlength{\tabcolsep}{0.05pt}
    \renewcommand{\arraystretch}{0.01}
    \resizebox{\linewidth}{!}{ %
        \begin{tabular}{ccccc}
              \includegraphics[width=2.89em, height=1.9em]{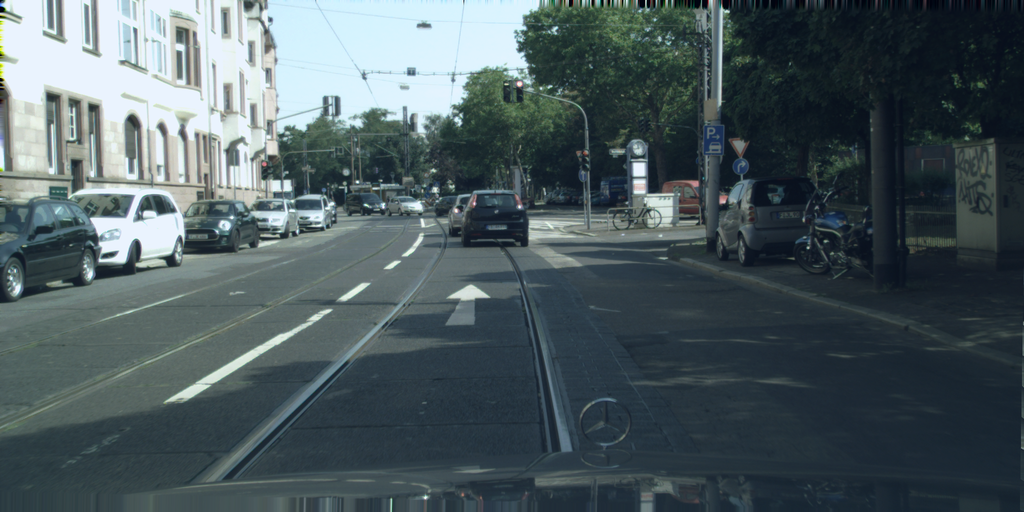} 
            & \includegraphics[width=2.89em, height=1.9em]{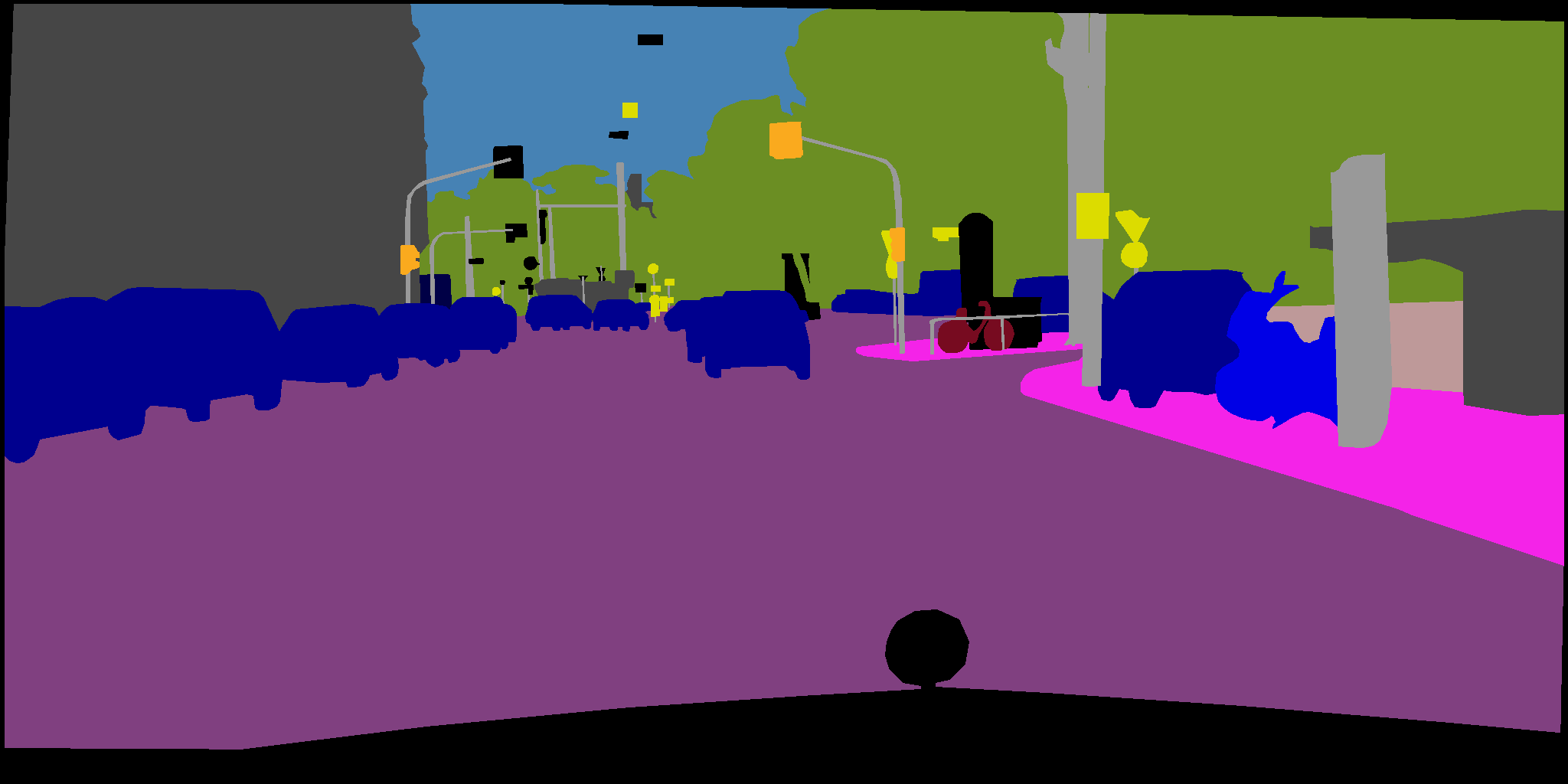}  
            & \includegraphics[width=2.89em, height=1.9em]{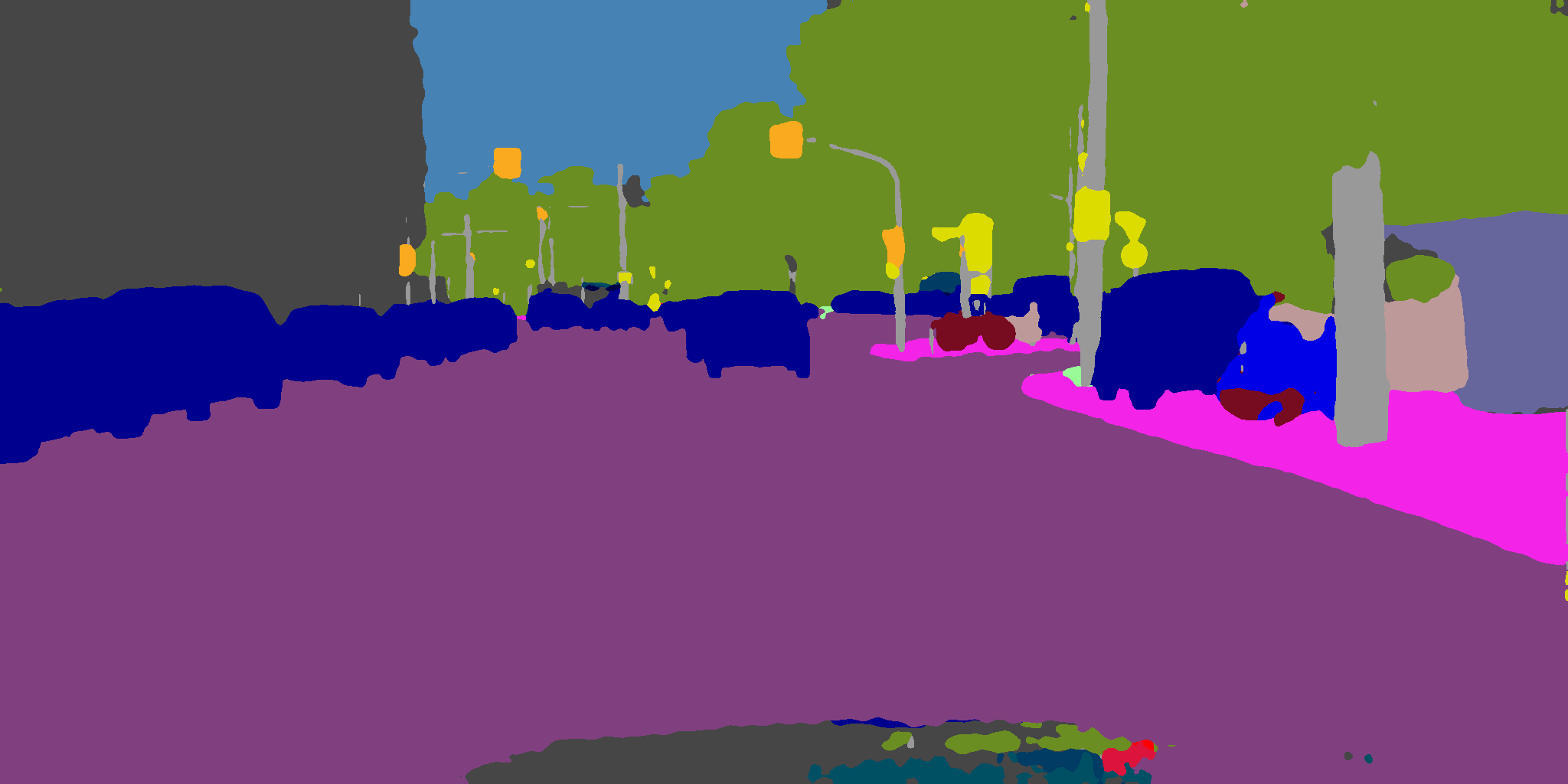} 
            & \includegraphics[width=2.89em, height=1.9em]{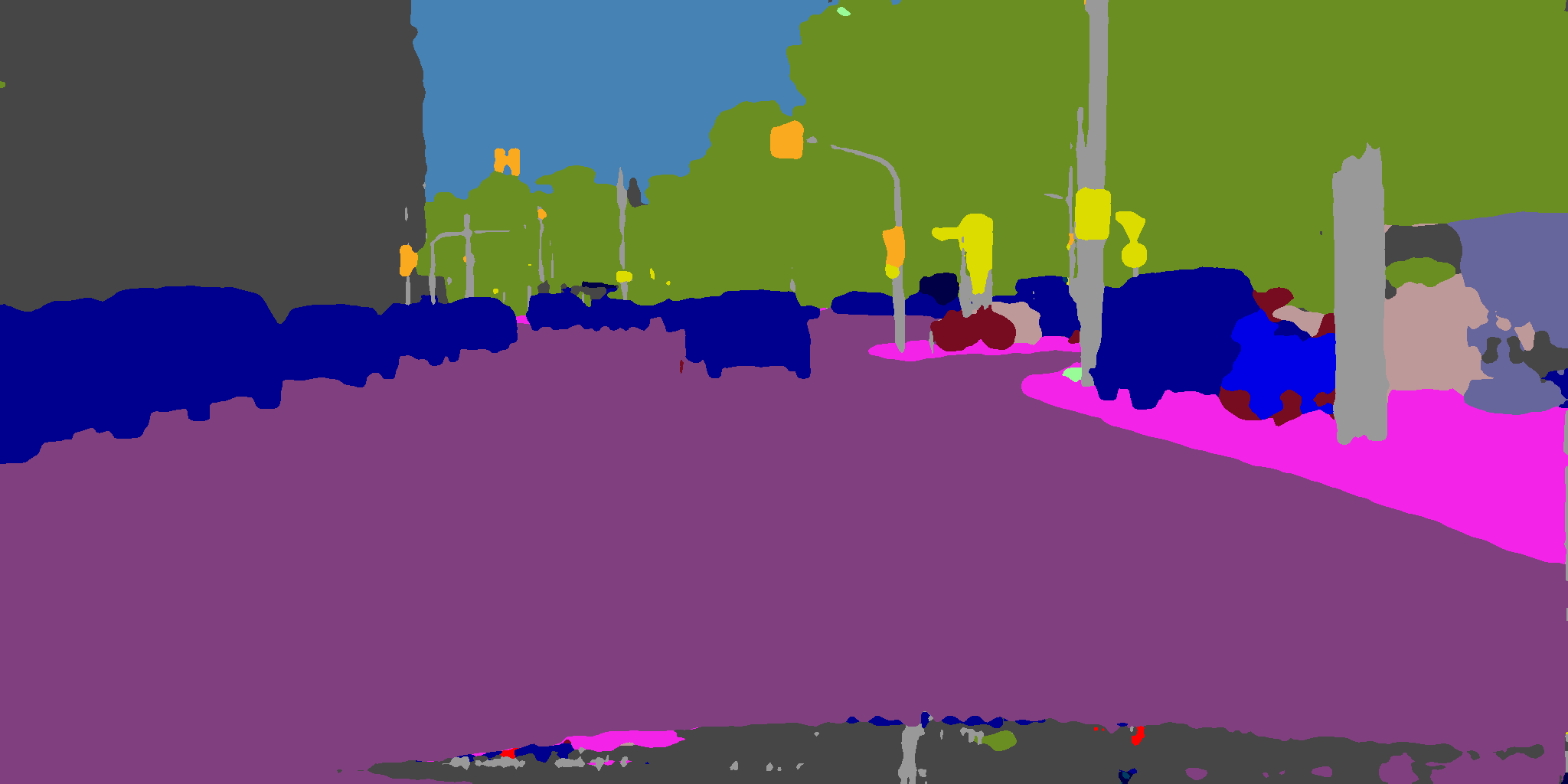}
            & \includegraphics[width=2.89em, height=1.9em]{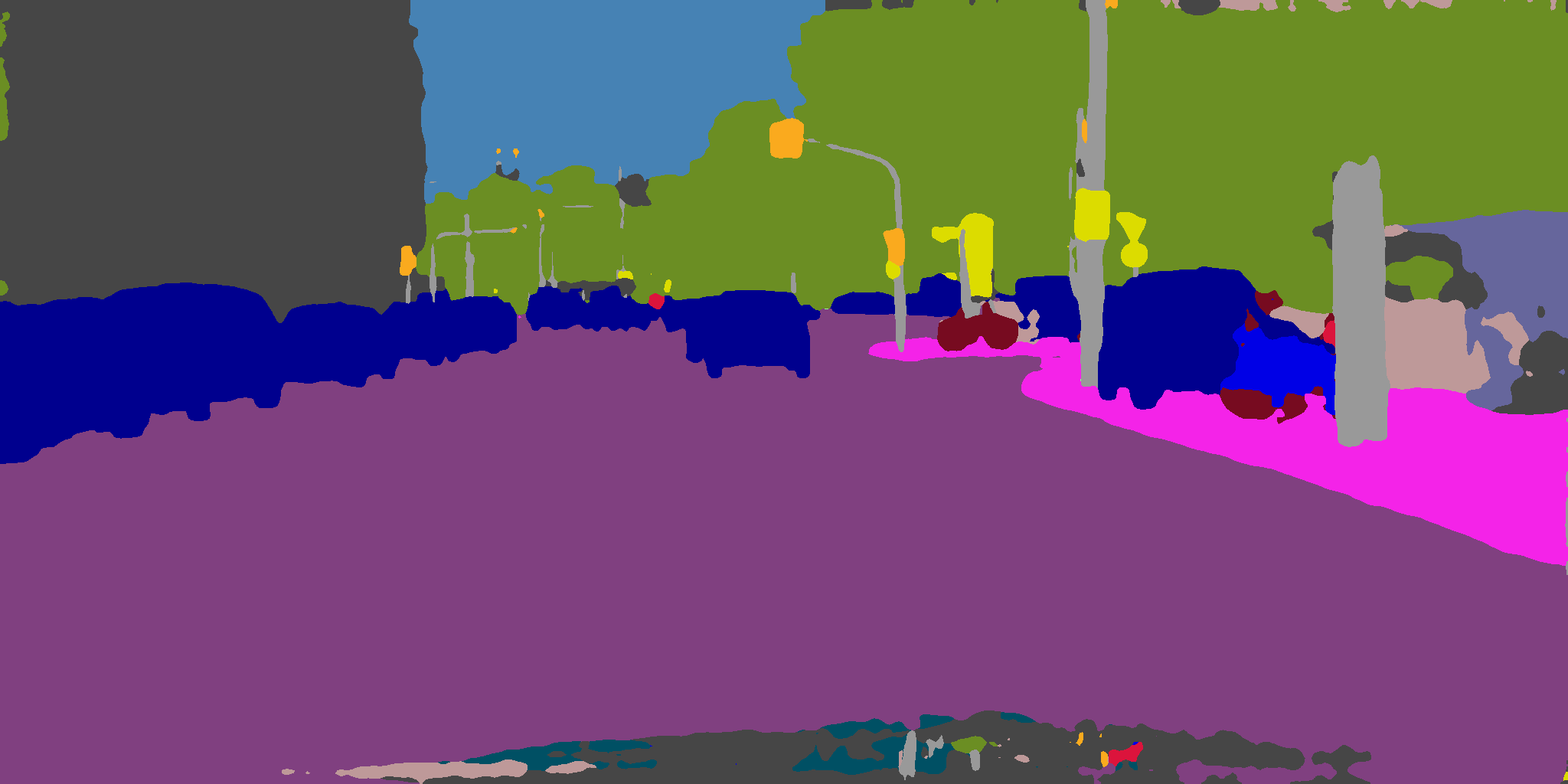}
            \\
              \includegraphics[width=2.89em, height=1.9em]{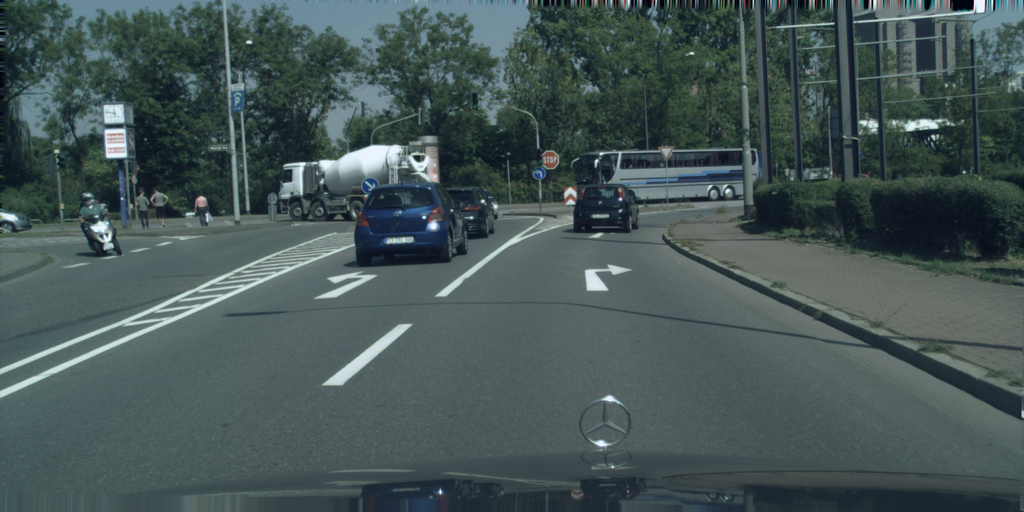} 
            & \includegraphics[width=2.89em, height=1.9em]{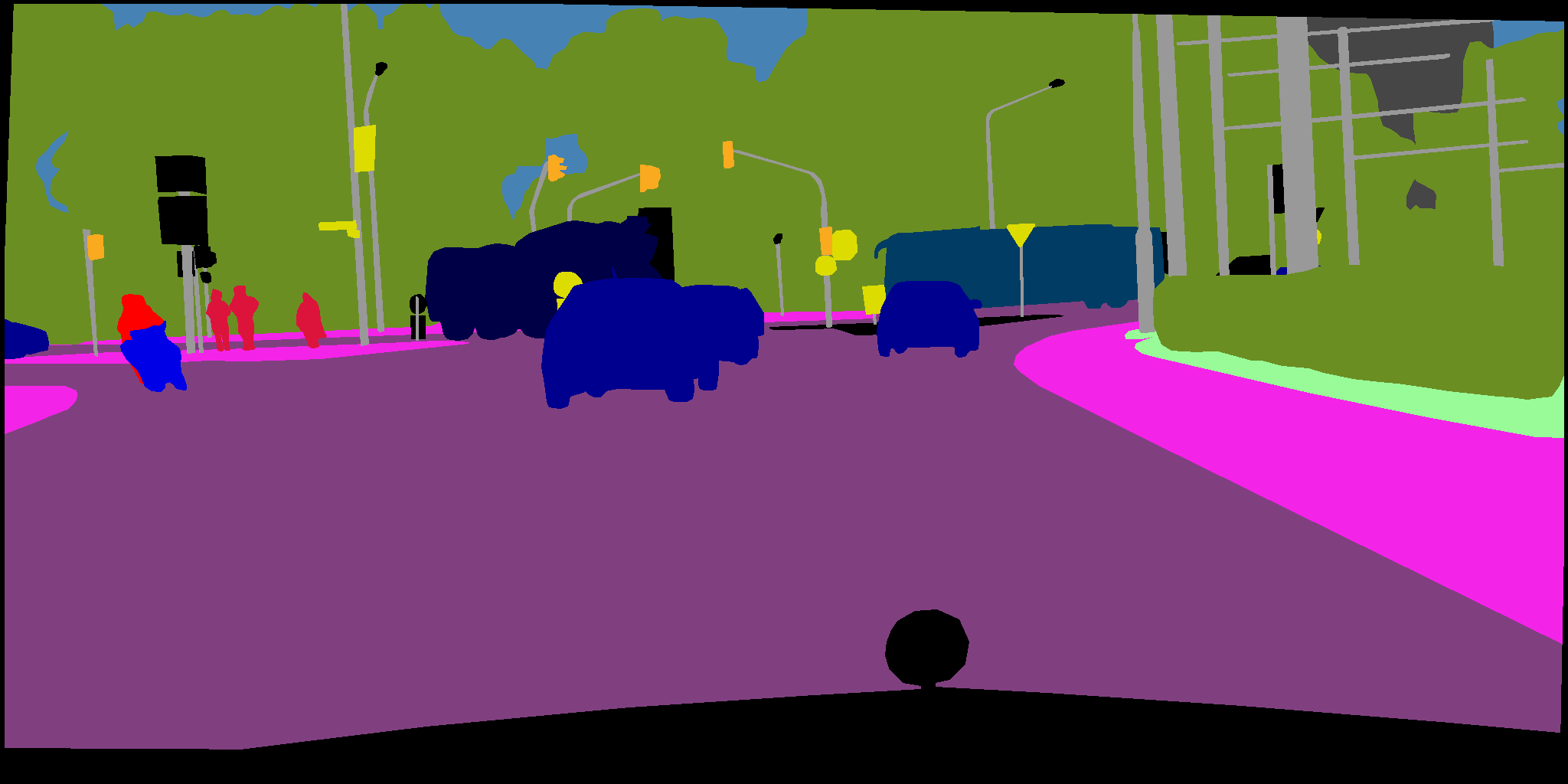}  
            & \includegraphics[width=2.89em, height=1.9em]{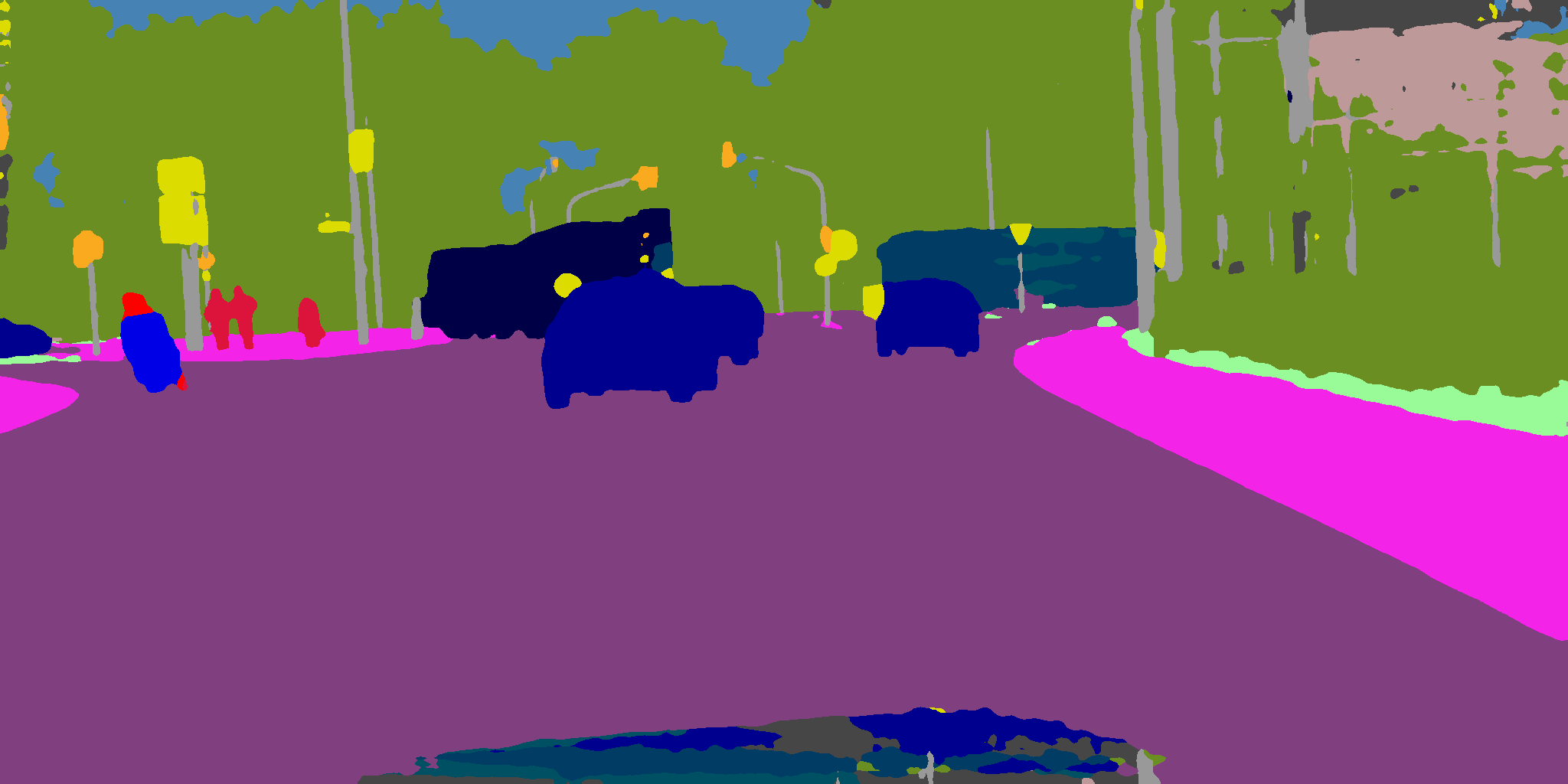} 
            & \includegraphics[width=2.89em, height=1.9em]{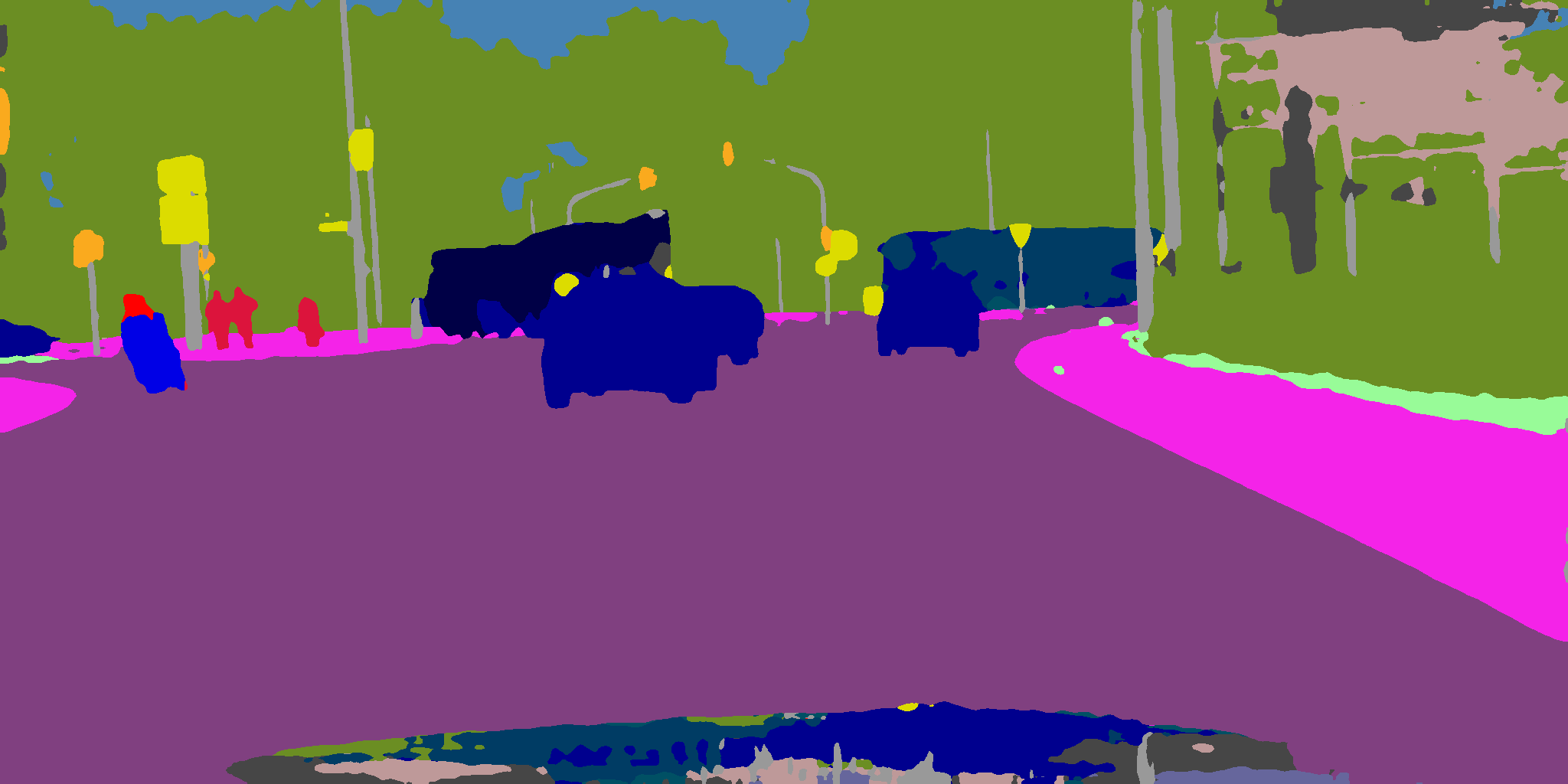}
            & \includegraphics[width=2.89em, height=1.9em]{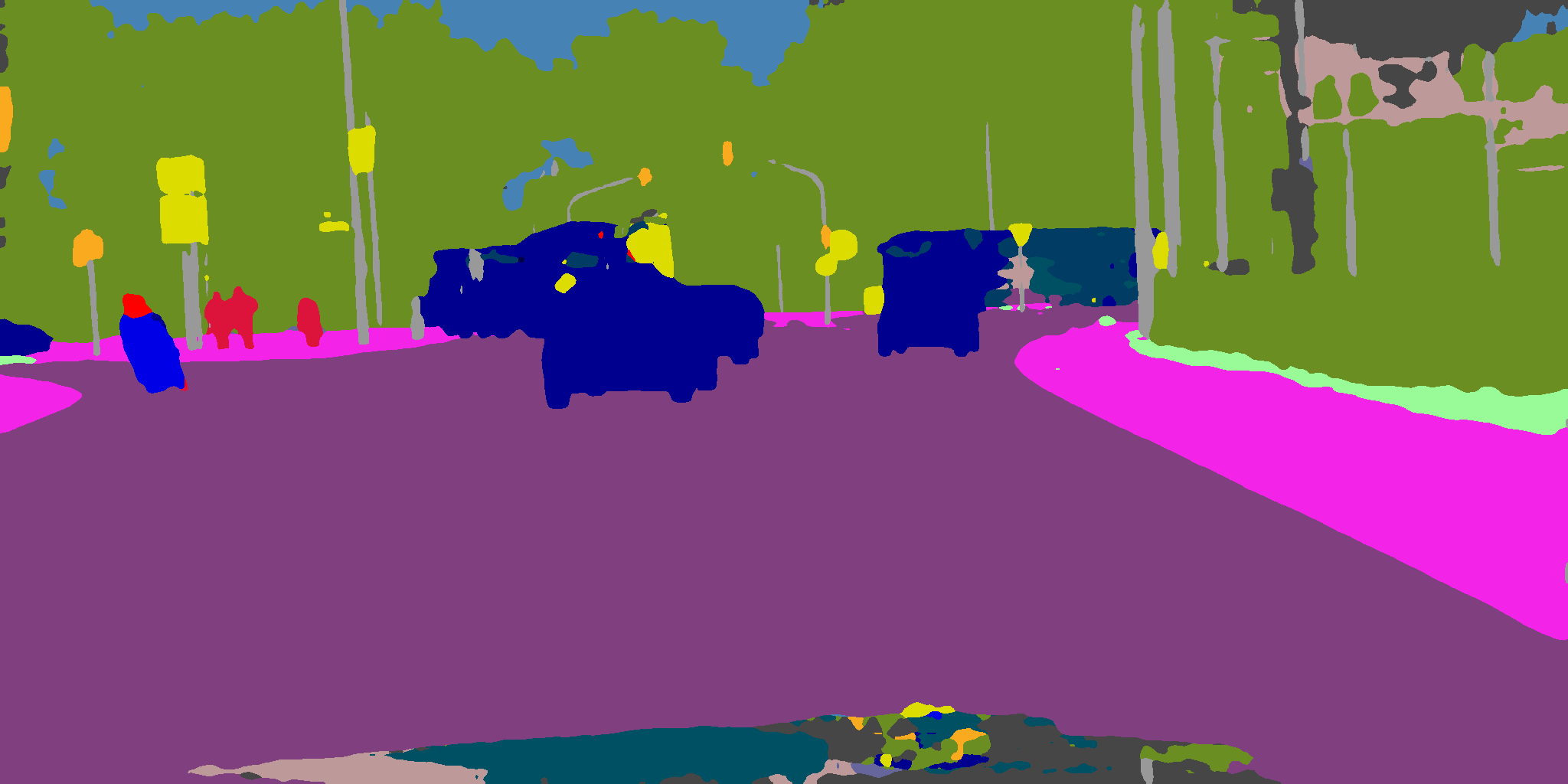}
            \\
              \includegraphics[width=2.89em, height=1.9em]{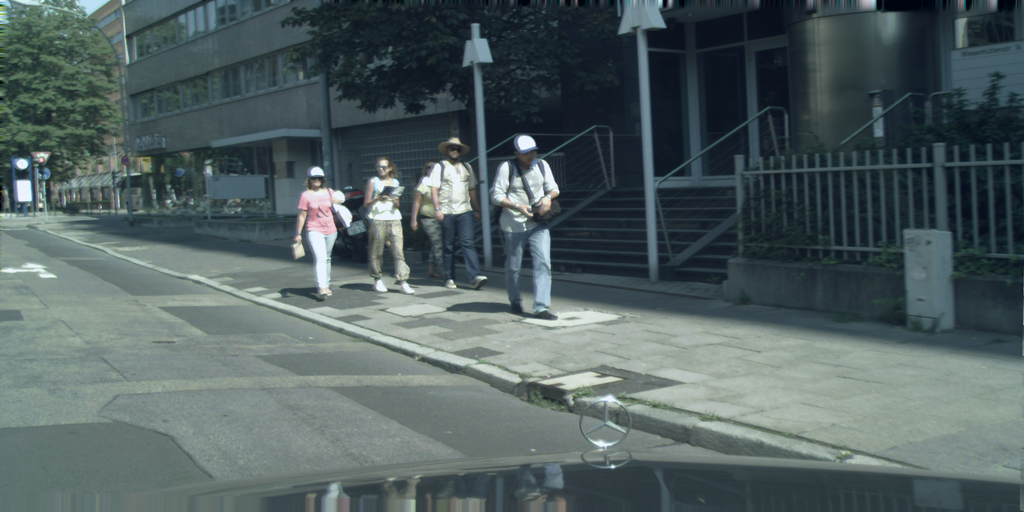} 
            & \includegraphics[width=2.89em, height=1.9em]{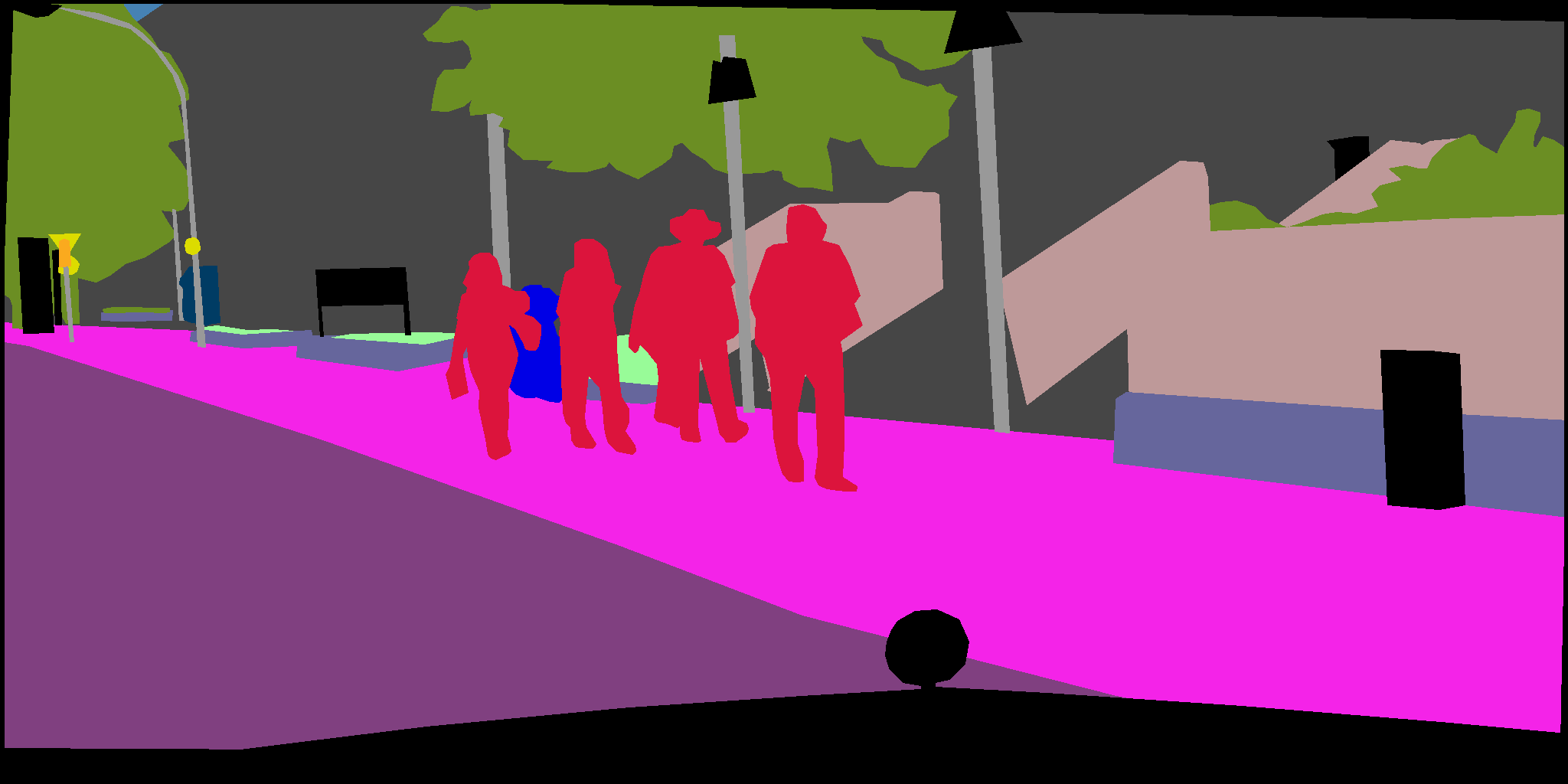}  
            & \includegraphics[width=2.89em, height=1.9em]{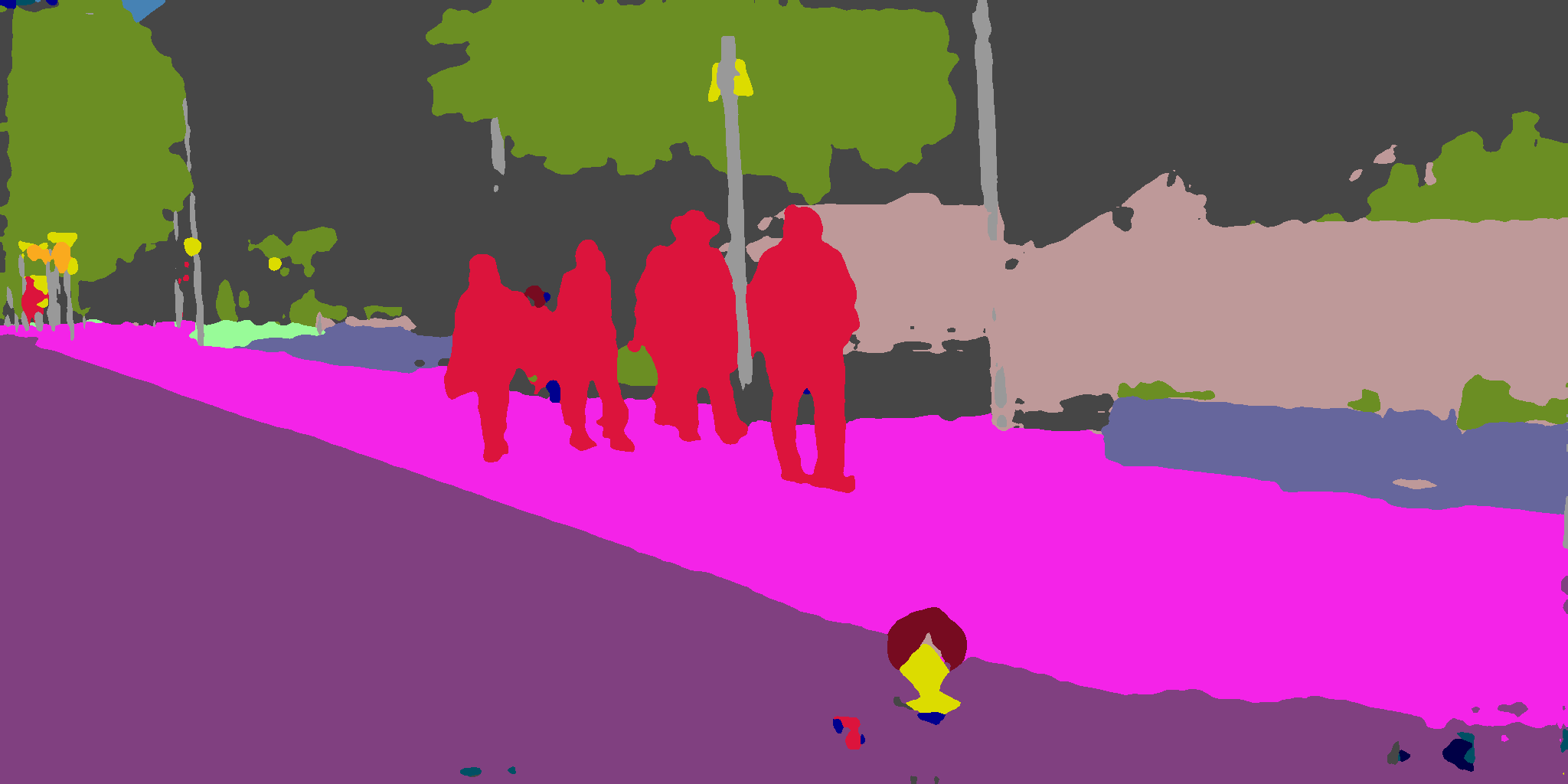} 
            & \includegraphics[width=2.89em, height=1.9em]{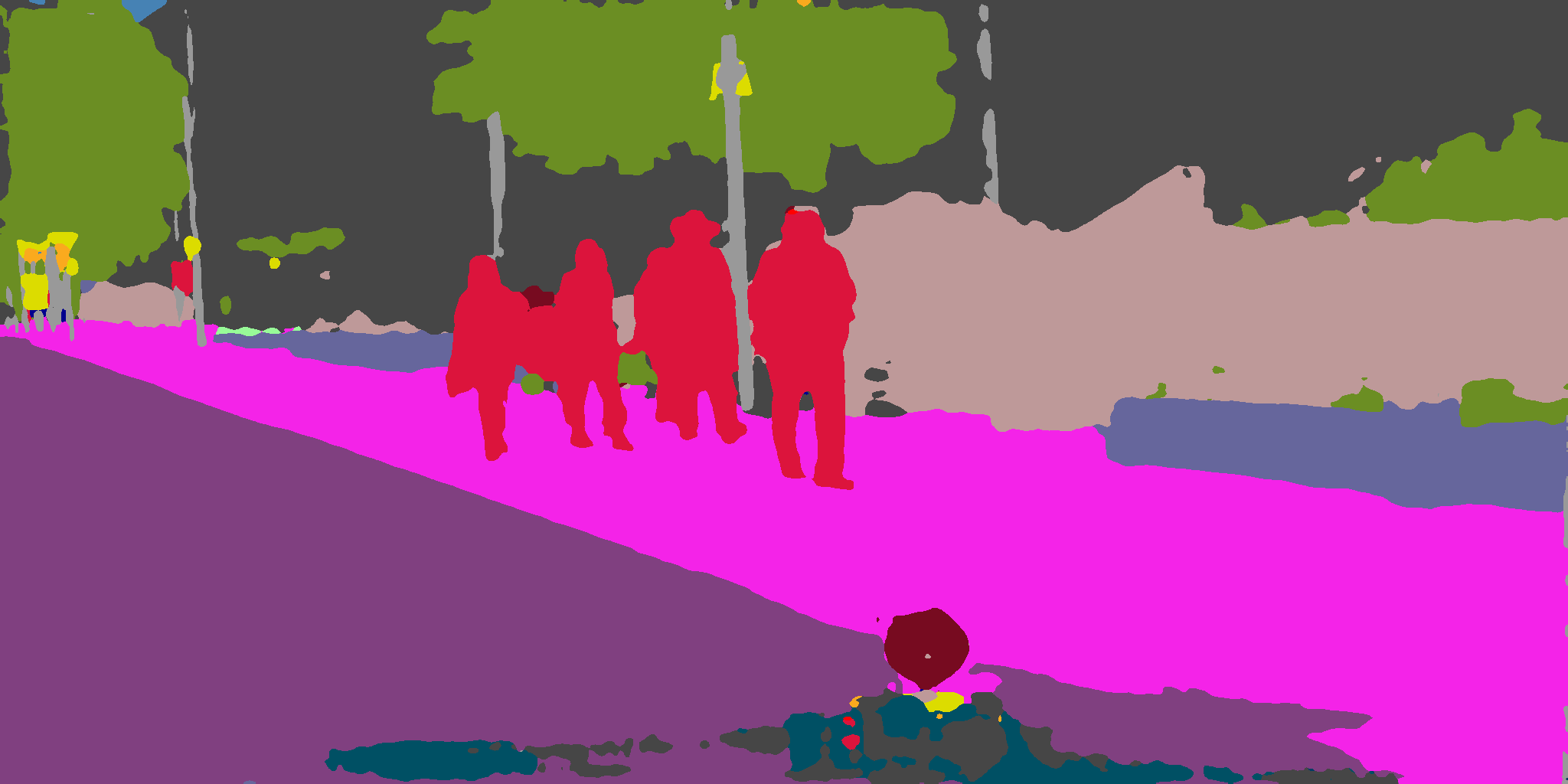}
            & \includegraphics[width=2.89em, height=1.9em]{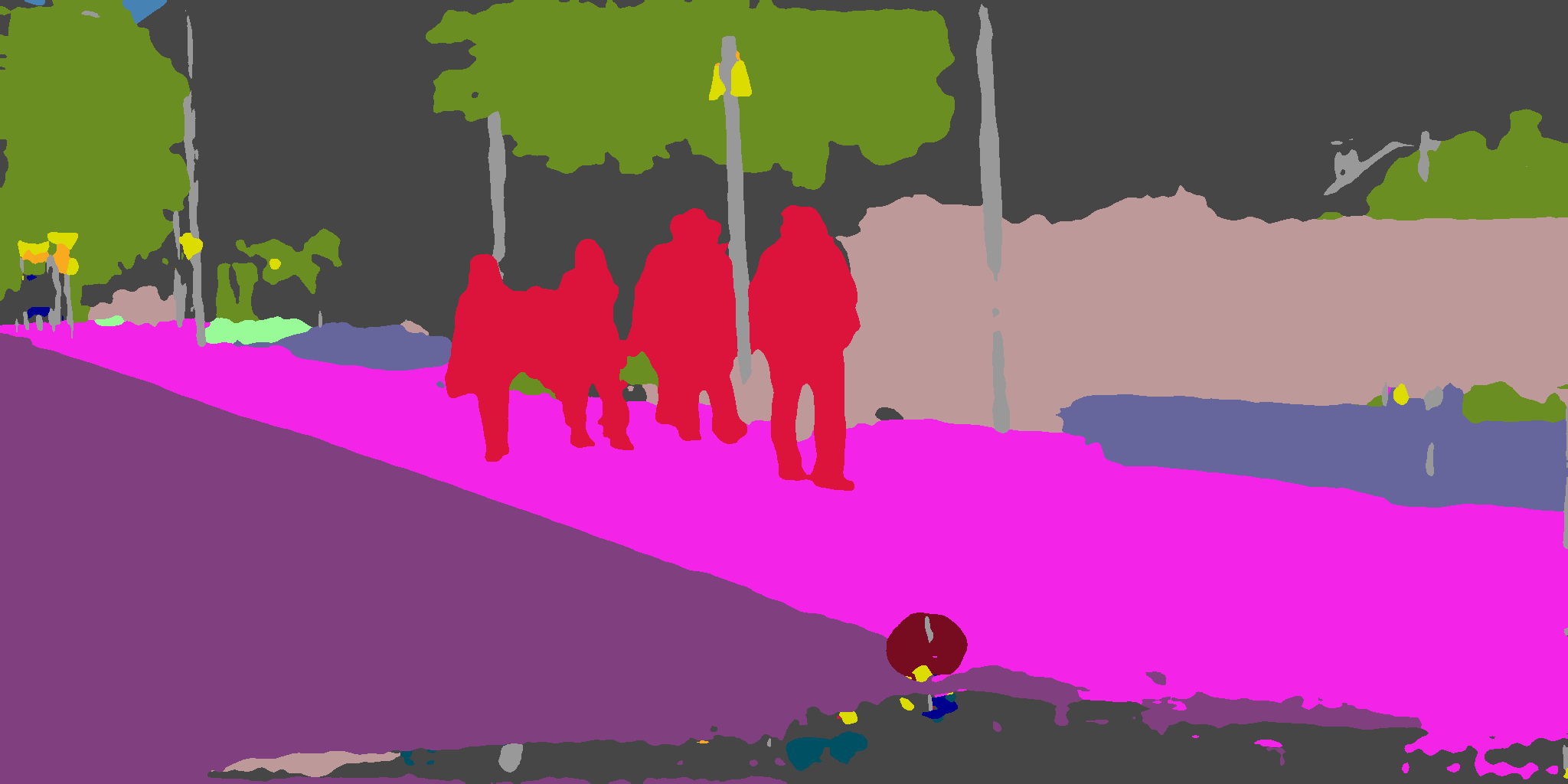}
            \\
            &&&&\\
            &&&&\\
            &&&&\\
            &&&&\\
              {\fontsize{3}{3.5}\selectfont Input Image}
            & {\fontsize{3}{3.5}\selectfont GT}
            & {\fontsize{3}{3.5}\selectfont 1/2}
            & {\fontsize{3}{3.5}\selectfont 1/4}
            & {\fontsize{3}{3.5}\selectfont 1/8}

        \end{tabular}
    }
    \caption{Qualitative results on a \emph{val} set of Cityscapes in various proportions of labeled data to unlabeled data.}
    \label{sec:qual-2-city}
\end{figure*}

\noindent\textbf{Performance analysis on semantic segmentation.}
To demonstrate the superior performance of our method, we compare the method with recent state-of-the-art models and training on labeled data only (Baseline). The results of our method on PASCAL VOC 2012 are listed in Table \hyperref[sec:VOC-RESULT]{1}. We abbreviate Deeplab v2 to DL2, Deeplab v3+ to DL3+ and ResNet-50 to R50, ResNet-101 to R101. To test the performance of our method under various conditions, we conduct experiments on three ratios (1/20, 1/8, 1/4) with ResNet-50 and ResNet-101 as a backbone network, respectively. As we can see from the table, our method achieves superior performance over all other works for both backbone networks. 
It is considered that our error localization concept is much more effective than error correction in the semi-supervised scheme from the comparison of results between ECS~\cite{ECNet} and Ours. We achieve higher performance than ECS with less labeled data; note that the performance of ECS is 70.22 in the 1/8 ratio, while Ours is 70.52 in the 1/20. Moreover, we conduct experiments on Cityscapes on three ratios (1/8, 1/4, 1/2) to show the generalization capability of our method. The results are displayed in Table \hyperref[sec:CITY-RESULT]{2}, showing that our method still outperforms other methods. Fig. \hyperref[sec:qual-2-voc]{3} and \hyperref[sec:qual-2-city]{4} show qualitative results of our method under various ratio conditions.

\begin{figure*}[t!]
\centering
    \setlength{\tabcolsep}{0.02pt}
    \renewcommand{\arraystretch}{0.01}
    \resizebox{\textwidth}{!}{ %
        \begin{tabular}{cccccc}
              \includegraphics[width=3.0em, height=2.80em]{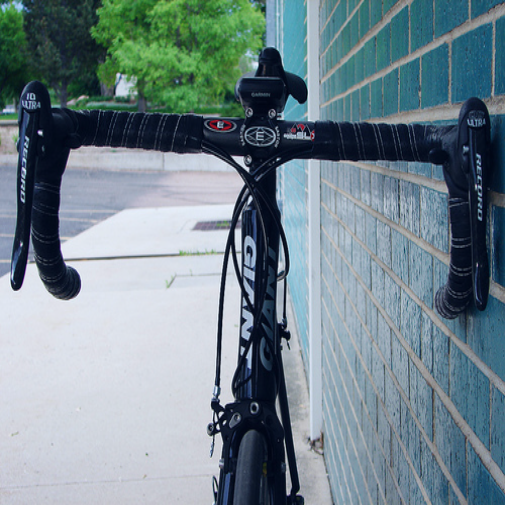} 
            & \includegraphics[width=3.0em, height=2.80em]{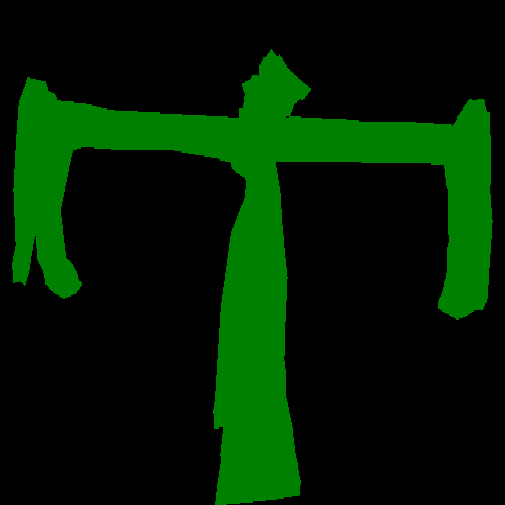}  
            & \includegraphics[width=3.0em, height=2.80em]{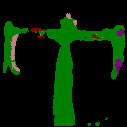}
            & \includegraphics[width=3.0em, height=2.80em]{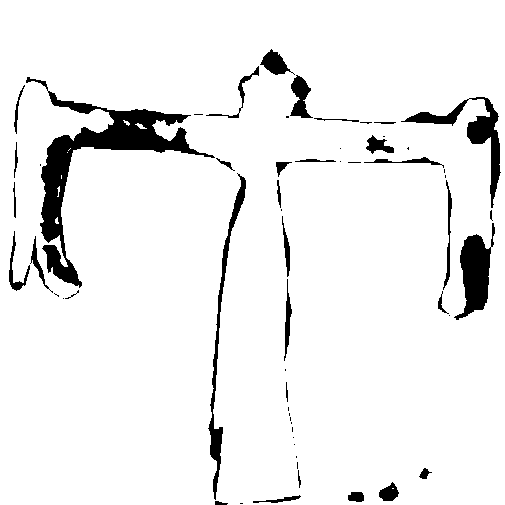} 
            & \includegraphics[width=3.0em, height=2.80em]{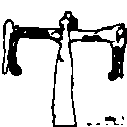} 
            
            & \includegraphics[width=3.0em, height=2.80em]{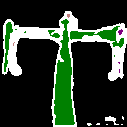}\\
 
              \includegraphics[width=3.0em, height=2.80em]{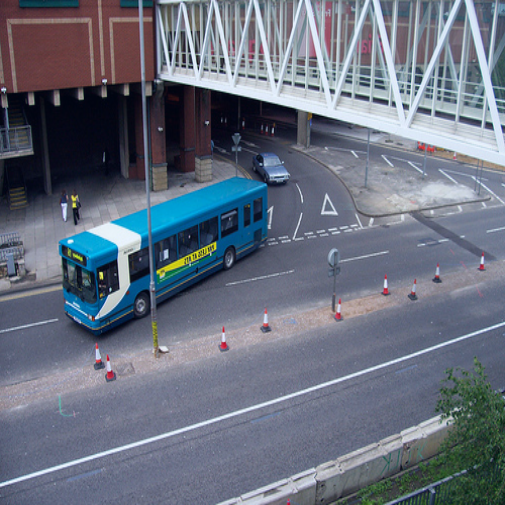} 
            & \includegraphics[width=3.0em, height=2.80em]{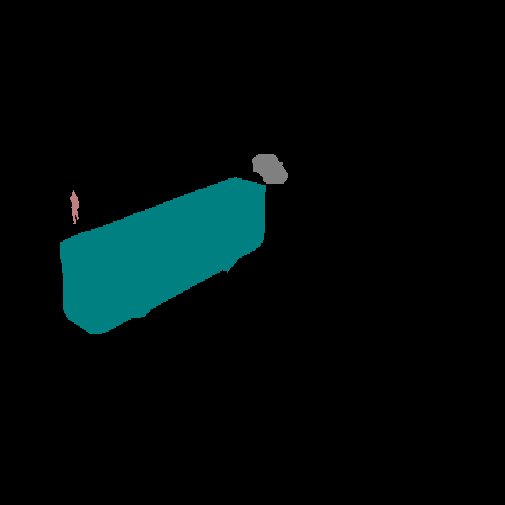}  
            & \includegraphics[width=3.0em, height=2.80em]{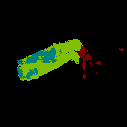} 
            & \includegraphics[width=3.0em, height=2.80em]{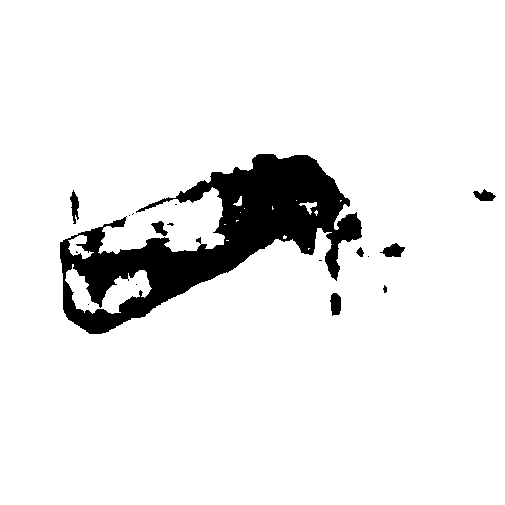}
            & \includegraphics[width=3.0em, height=2.80em]{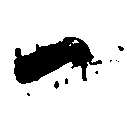} 
             
            & \includegraphics[width=3.0em, height=2.80em]{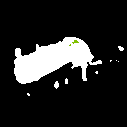}
            \\
              \includegraphics[width=3.0em, height=2.80em]{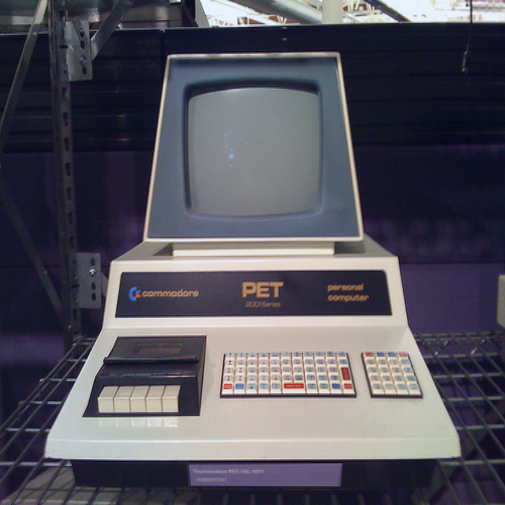} 
            & \includegraphics[width=3.0em, height=2.80em]{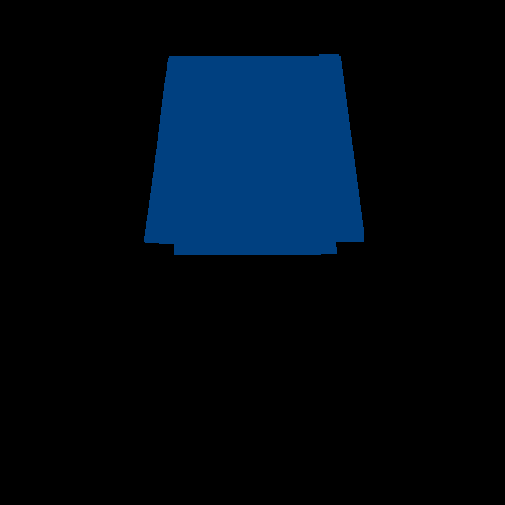}  
            & \includegraphics[width=3.0em, height=2.80em]{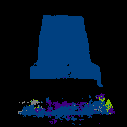} 
            & \includegraphics[width=3.0em, height=2.80em]{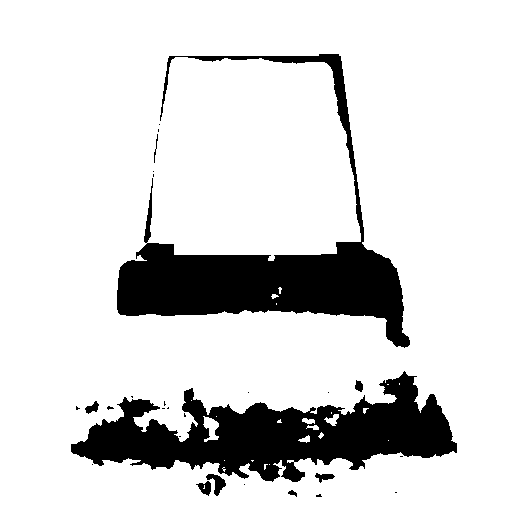} 
            & \includegraphics[width=3.0em, height=2.80em]{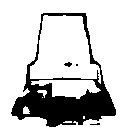} 
            & \includegraphics[width=3.0em, height=2.80em]{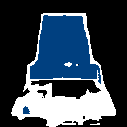}
            \\
            &&&&&\\
            &&&&&\\
              {\fontsize{3.5}{4.5}\selectfont Input Image}
            & {\fontsize{3.5}{4.5}\selectfont GT}
            & {\fontsize{3.5}{4.5}\selectfont (a)}
            & {\fontsize{3.5}{4.5}\selectfont (b)}
            & {\fontsize{3.5}{4.5}\selectfont (c)}
            & {\fontsize{3.5}{4.5}\selectfont (d)}
              
        \end{tabular}
    }
    \caption{Qualitative results on unlabeled data of training set on PASCAL VOC 2012 in the labeled ratio of 1/20. (a) Segmentation prediction from the main segmentation network. (b) Ground truth binary mask. (c) Binary mask predicted by ELN. (d) Filtered segmentation prediction by the predicted binary mask. Erroneous predictions colored in white in (d) are not used as pseudo labels.}
    \label{sec:qual-1-voc}
\end{figure*}

\begin{figure*}[t!]
\centering
    \setlength{\tabcolsep}{0.05pt}
    \renewcommand{\arraystretch}{0.01}
    \resizebox{\textwidth}{!}{ %
        \begin{tabular}{cccccc}
            
              \includegraphics[width=2.89em, height=1.9em]{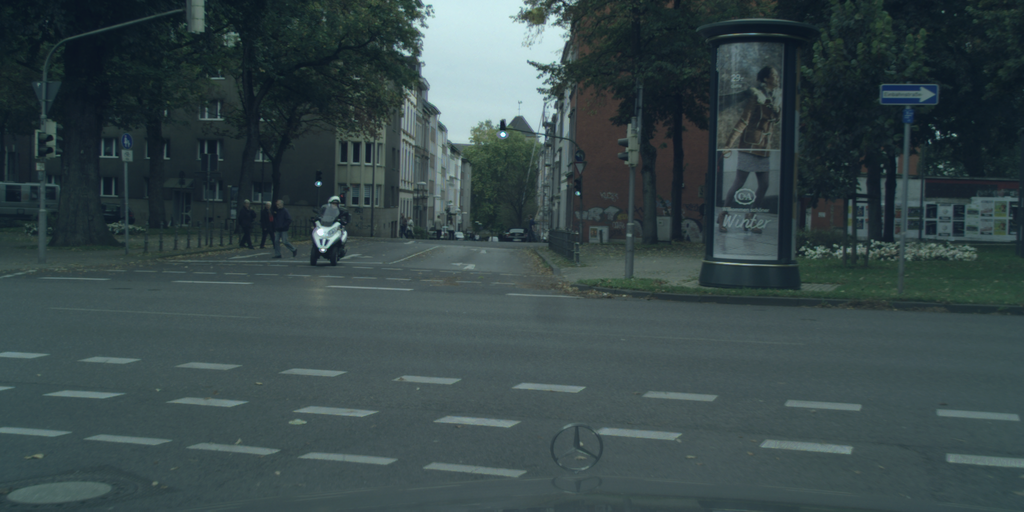} 
            & \includegraphics[width=2.89em, height=1.9em]{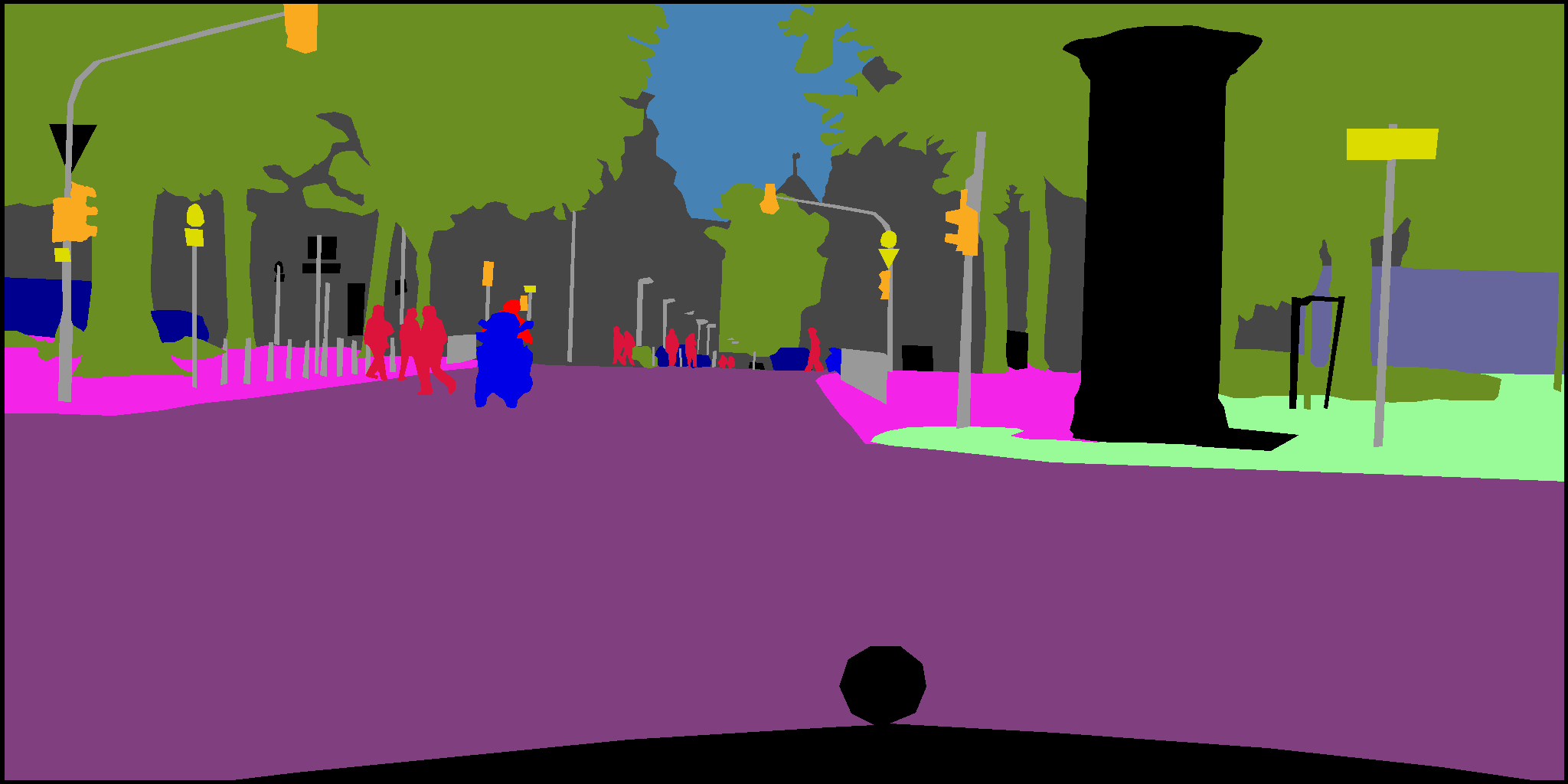}  
            & \includegraphics[width=2.89em, height=1.9em]{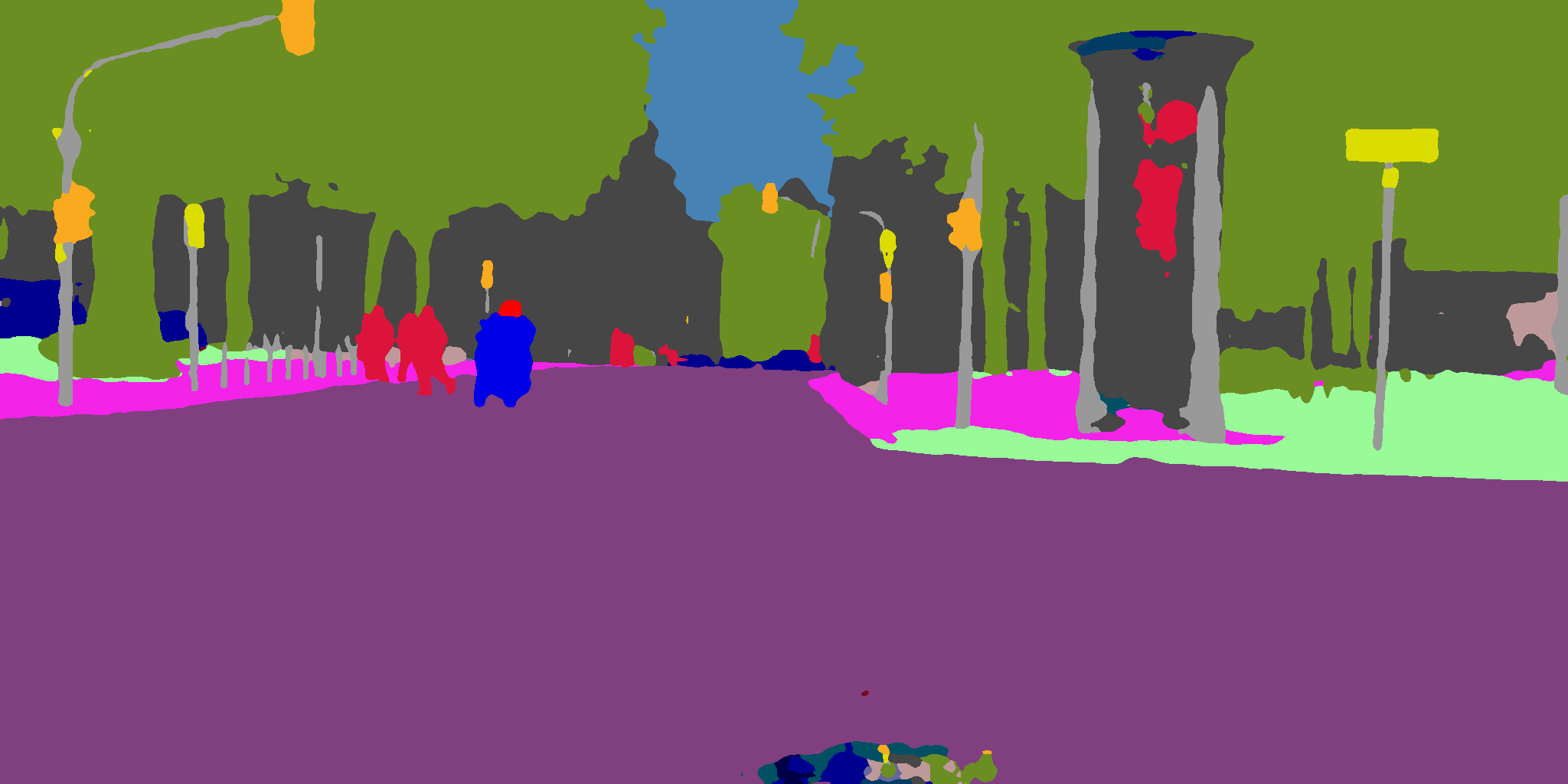}
            & \includegraphics[width=2.89em, height=1.9em]{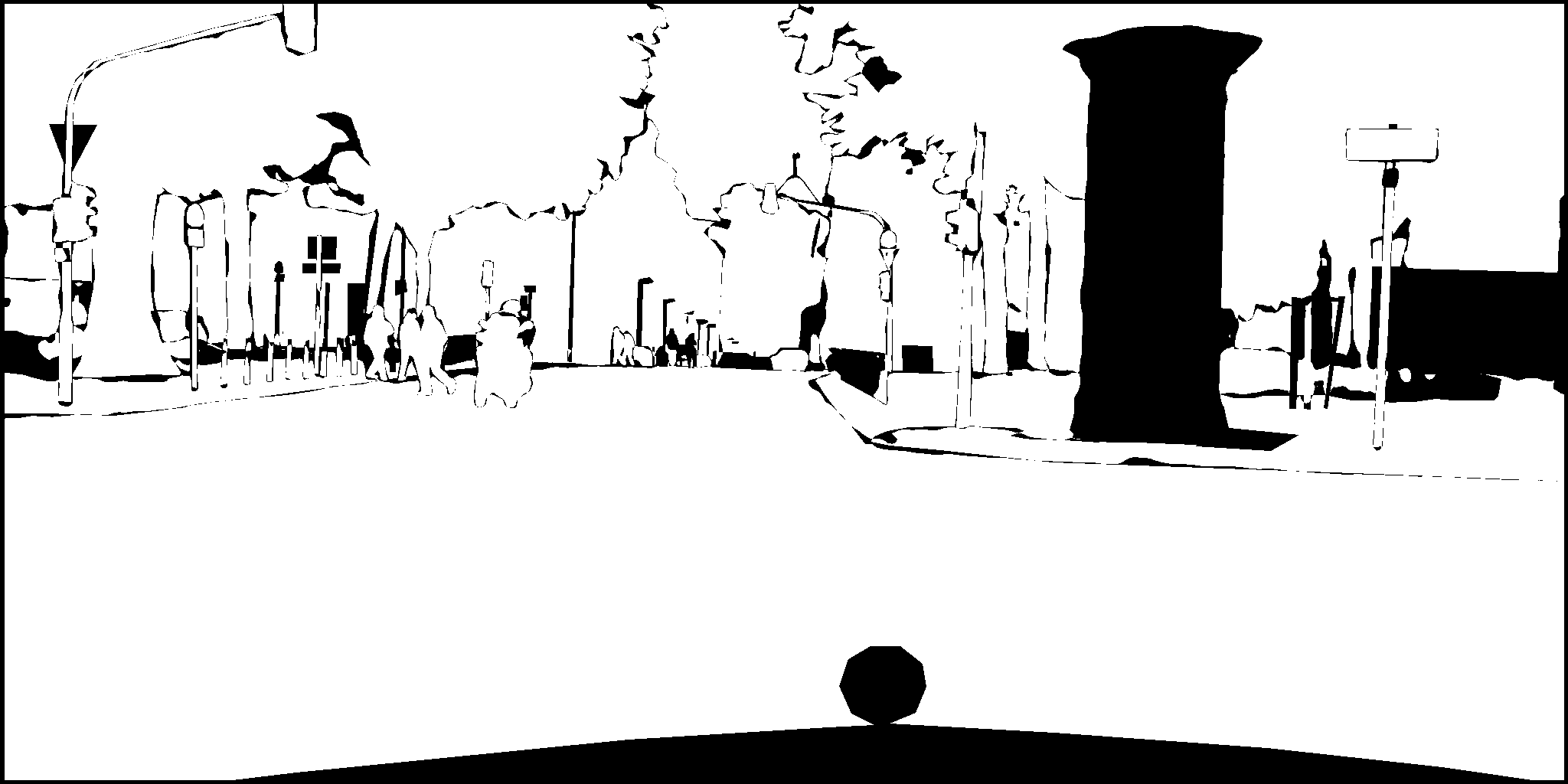} 
            & \includegraphics[width=2.89em, height=1.9em]{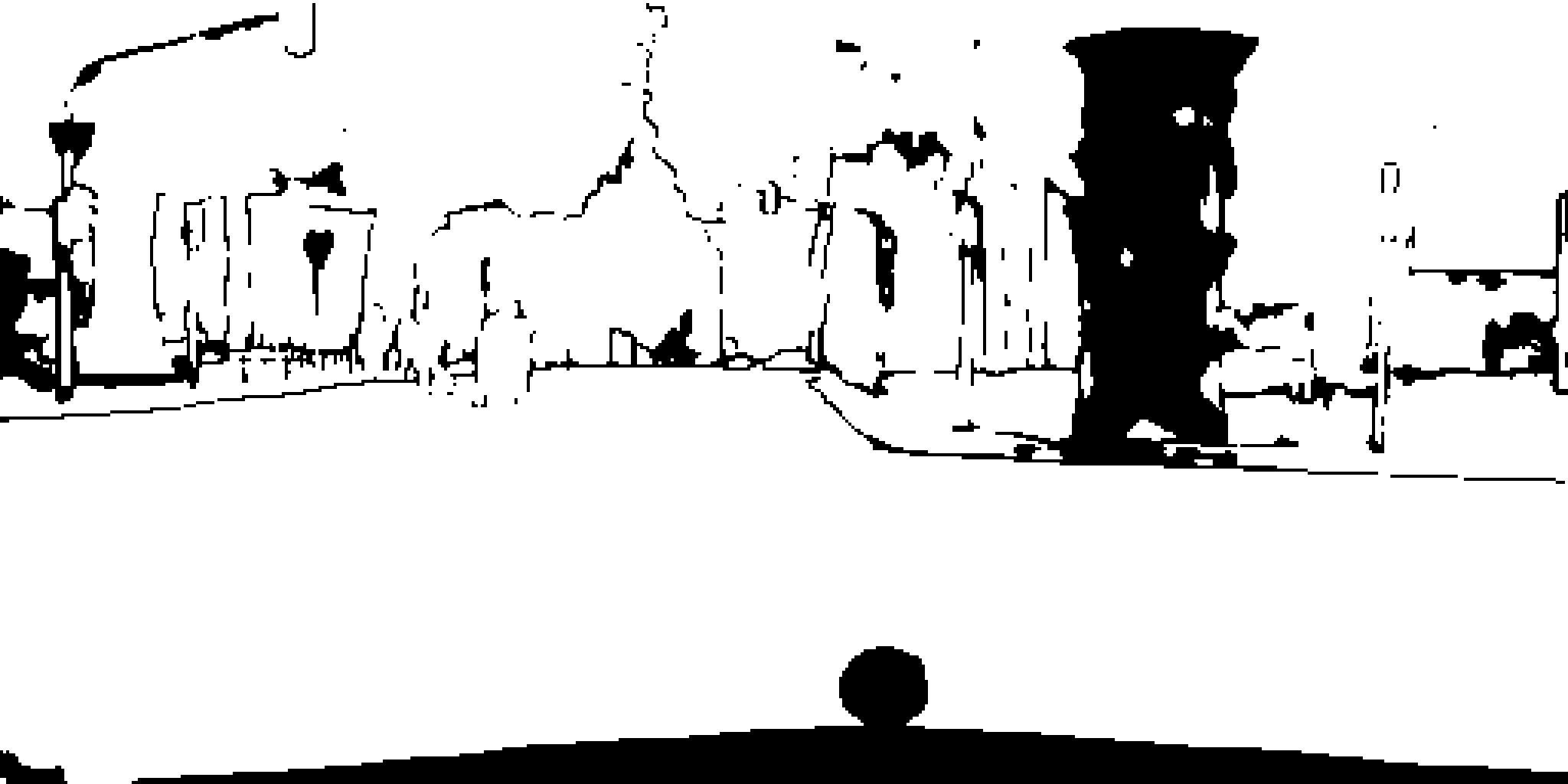}
            & \includegraphics[width=2.89em, height=1.9em]{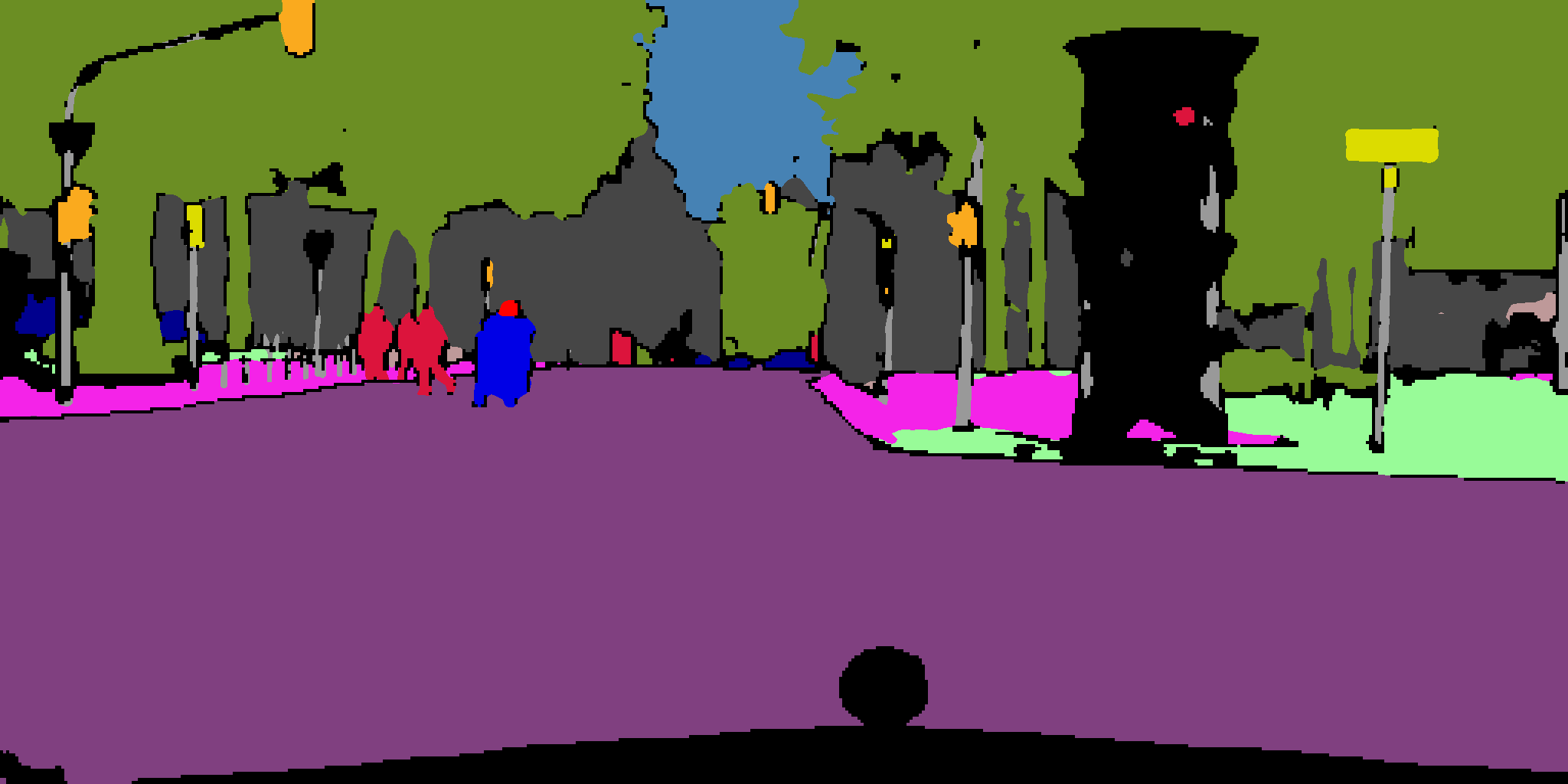}
            \\
              \includegraphics[width=2.89em, height=1.9em]{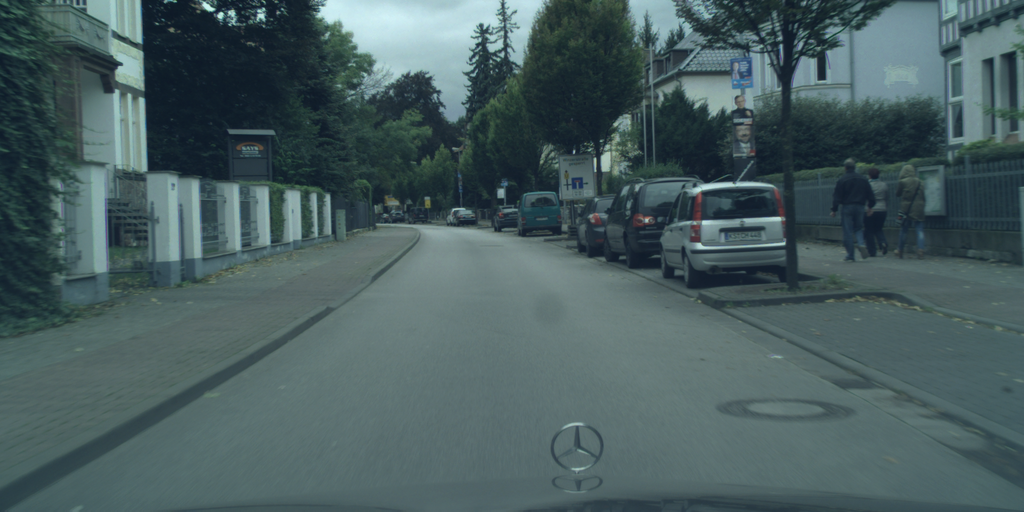} 
            & \includegraphics[width=2.89em, height=1.9em]{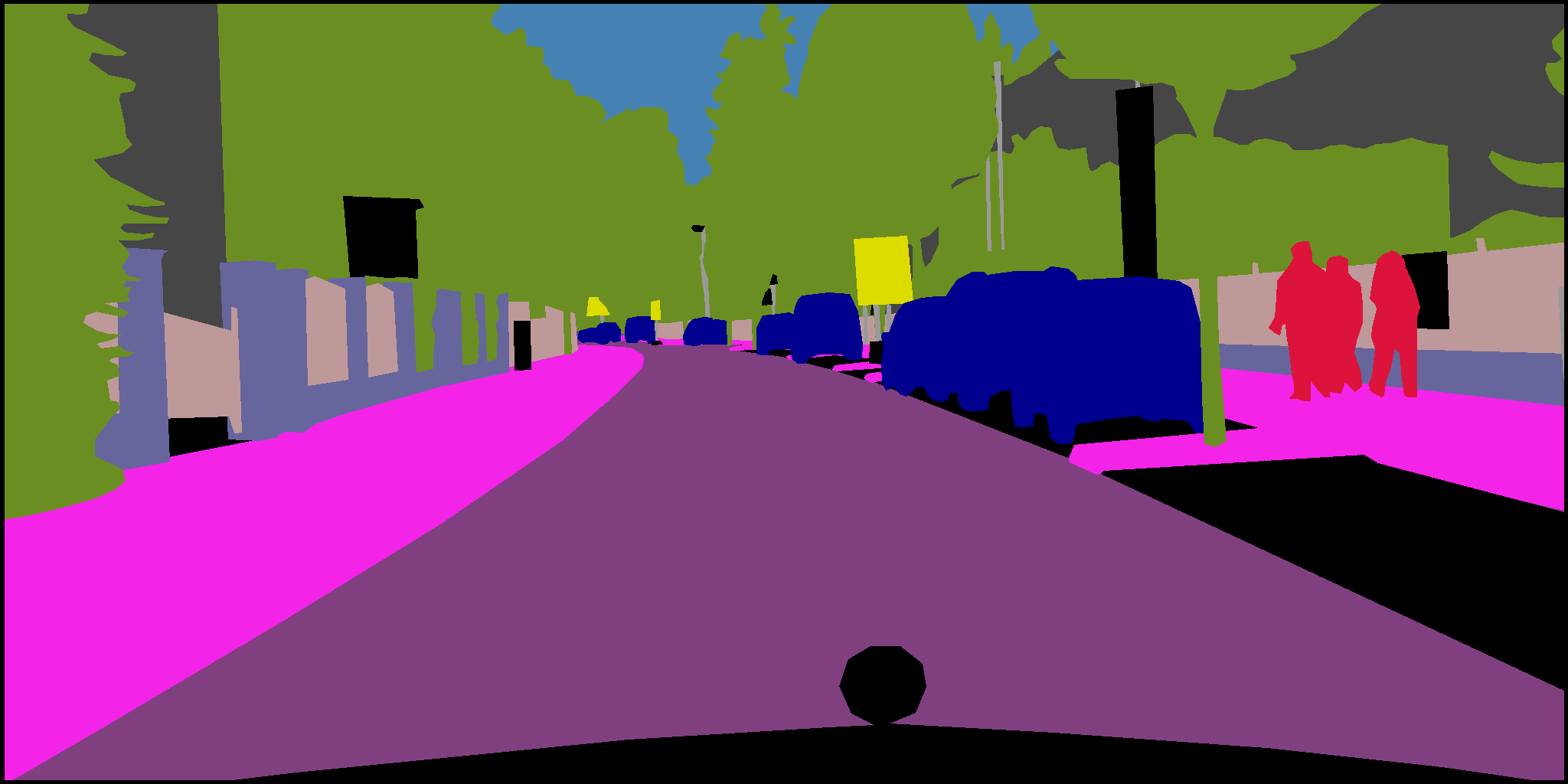}  
            & \includegraphics[width=2.89em, height=1.9em]{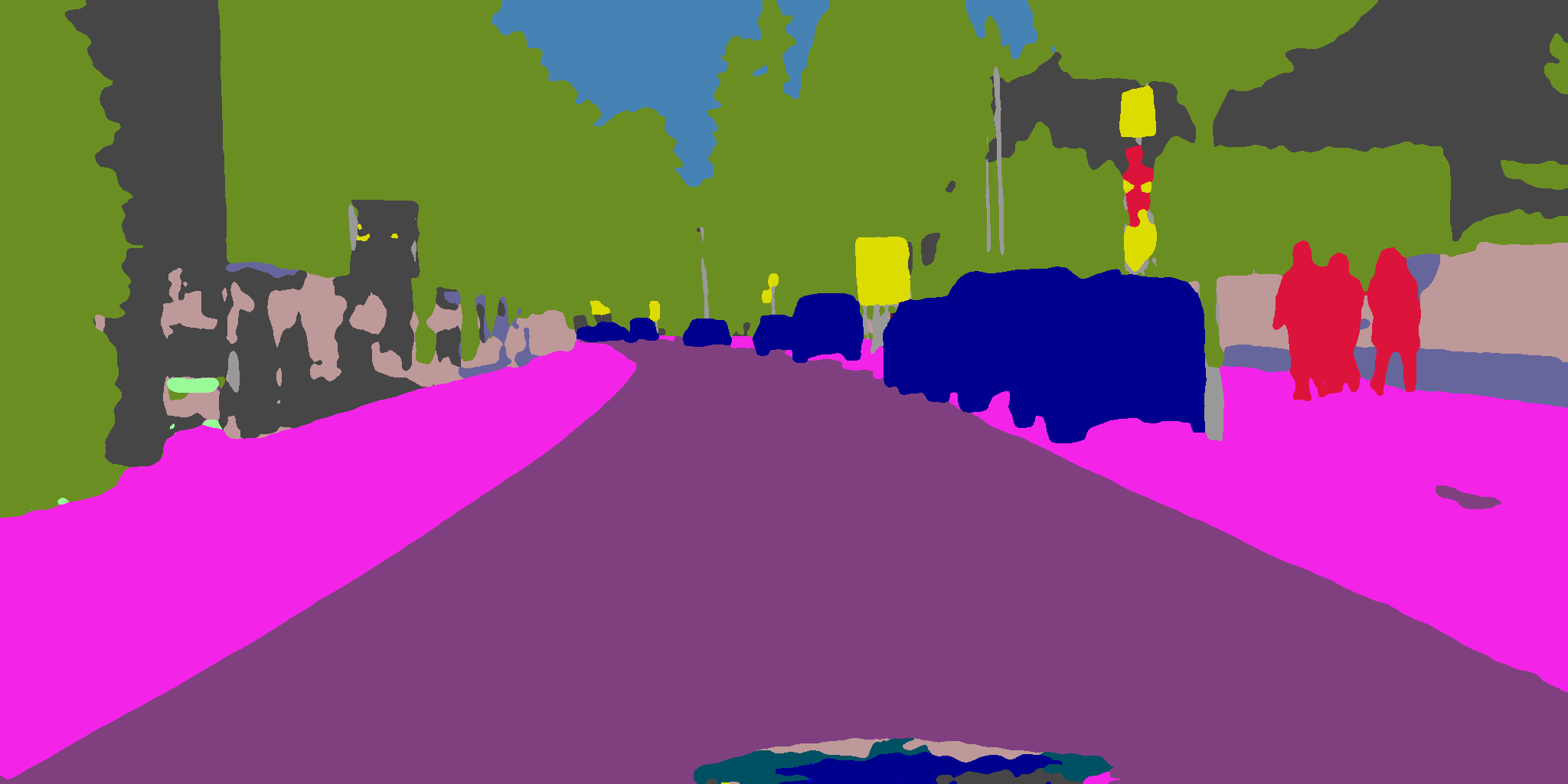}
            & \includegraphics[width=2.89em, height=1.9em]{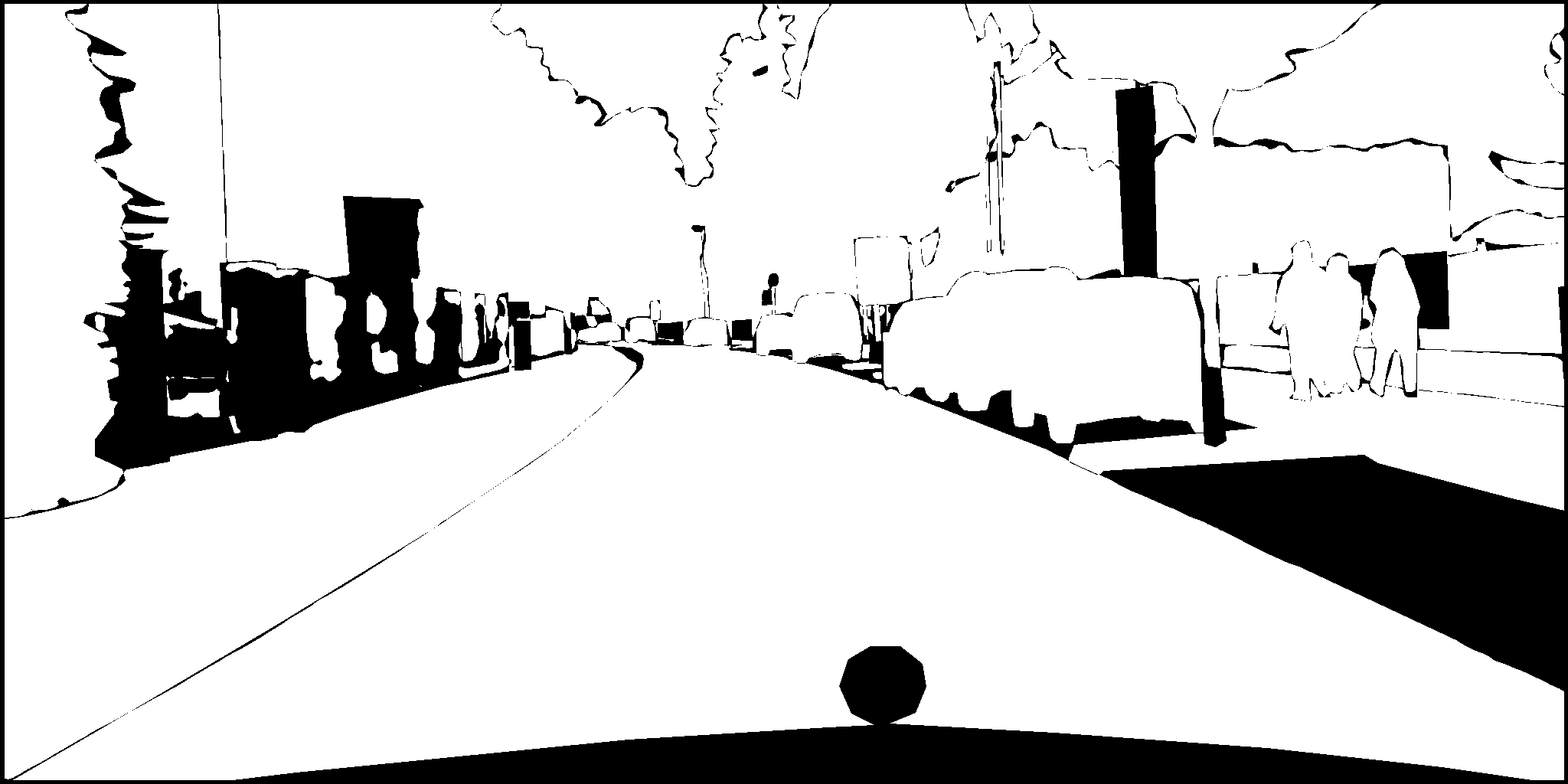} 
            & \includegraphics[width=2.89em, height=1.9em]{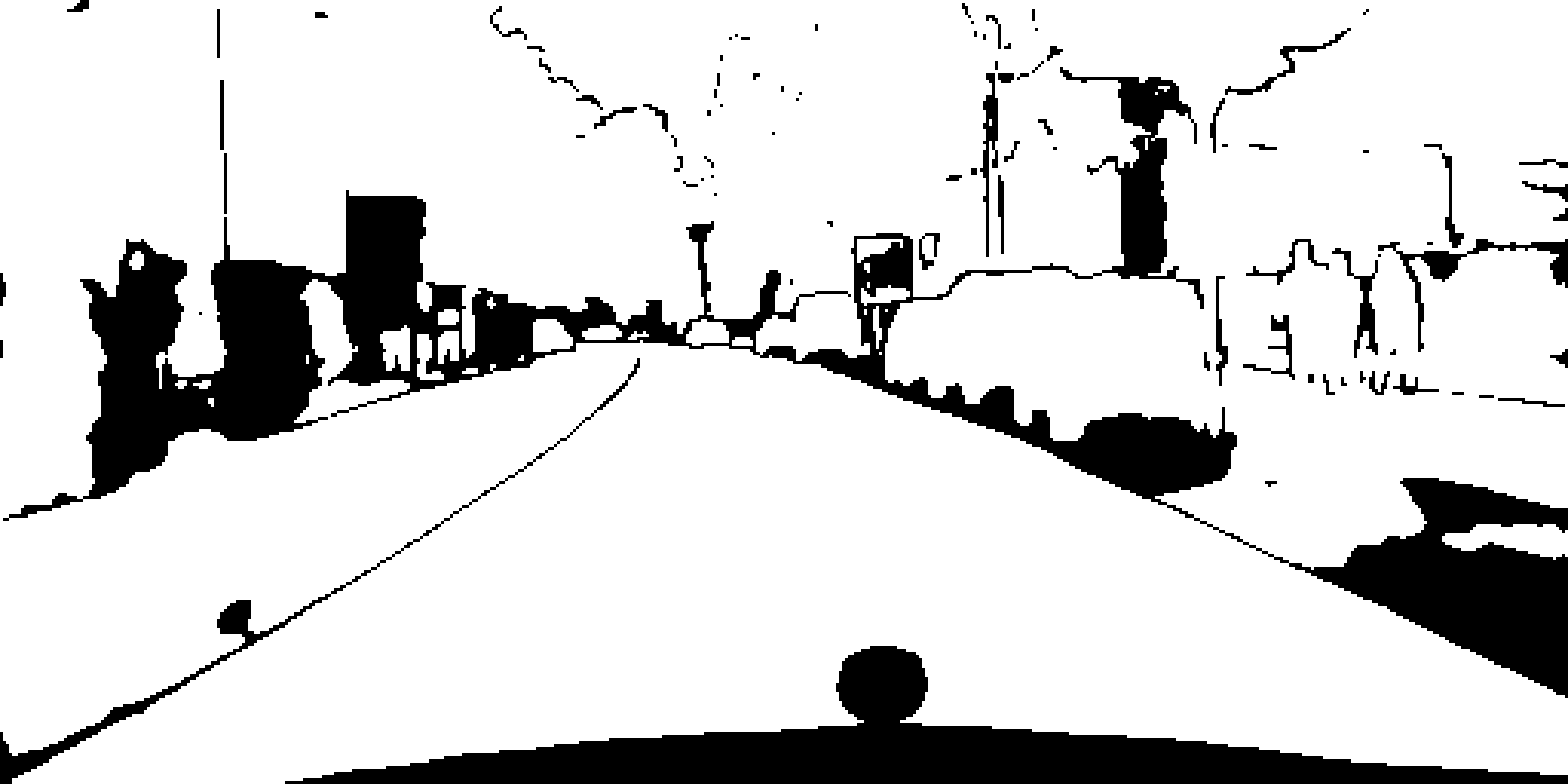}
            & \includegraphics[width=2.89em, height=1.9em]{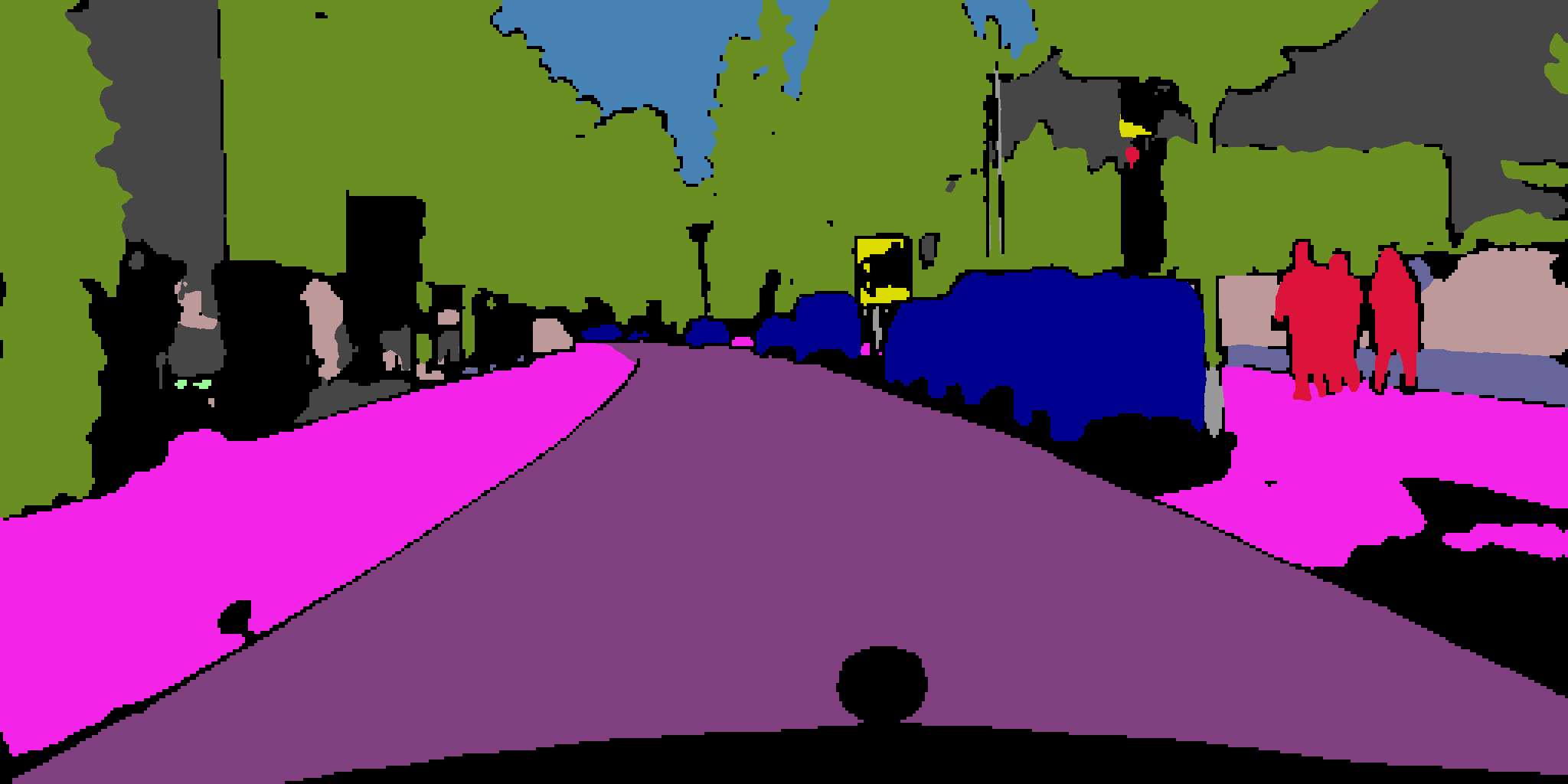}
            \\
              \includegraphics[width=2.89em, height=1.9em]{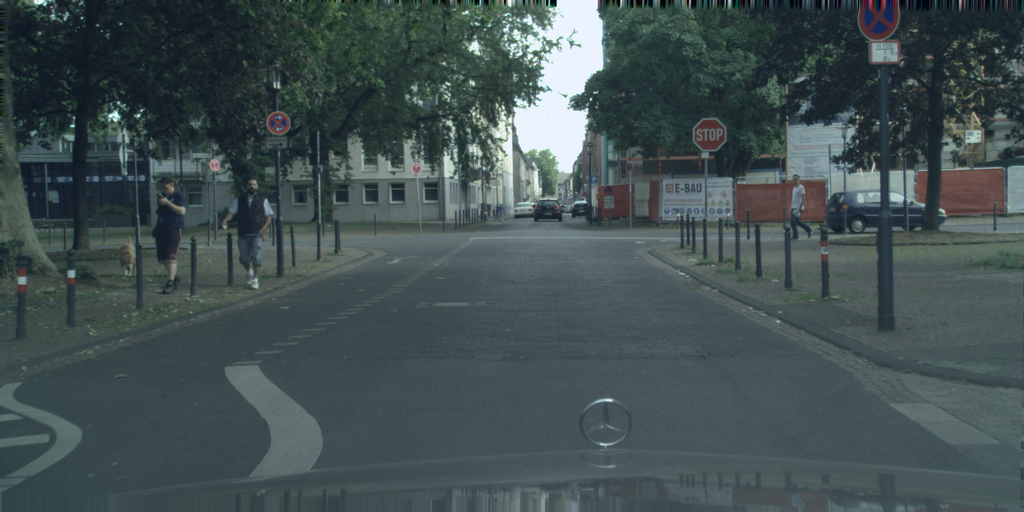} 
            & \includegraphics[width=2.89em, height=1.9em]{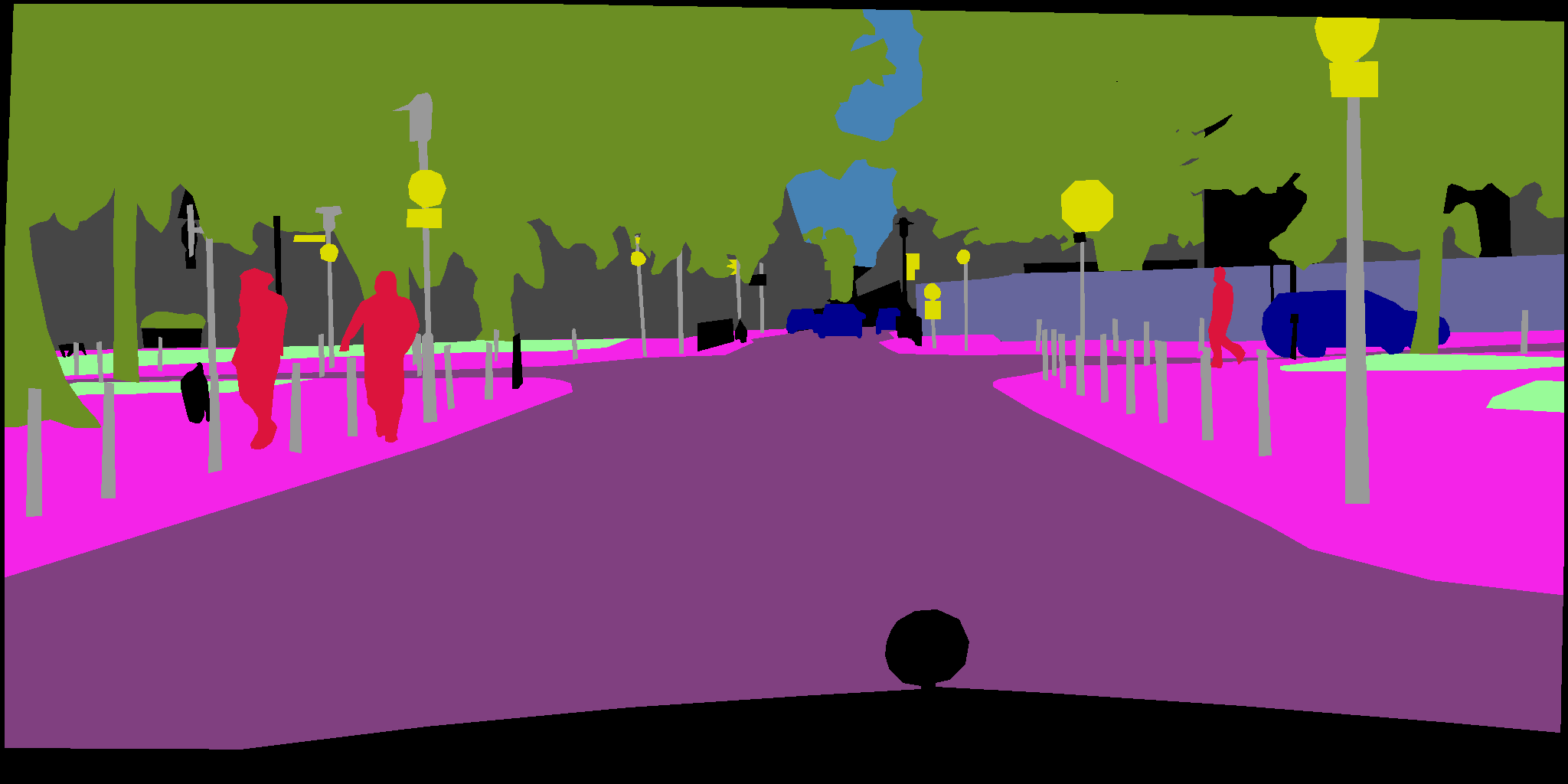}  
            & \includegraphics[width=2.89em, height=1.9em]{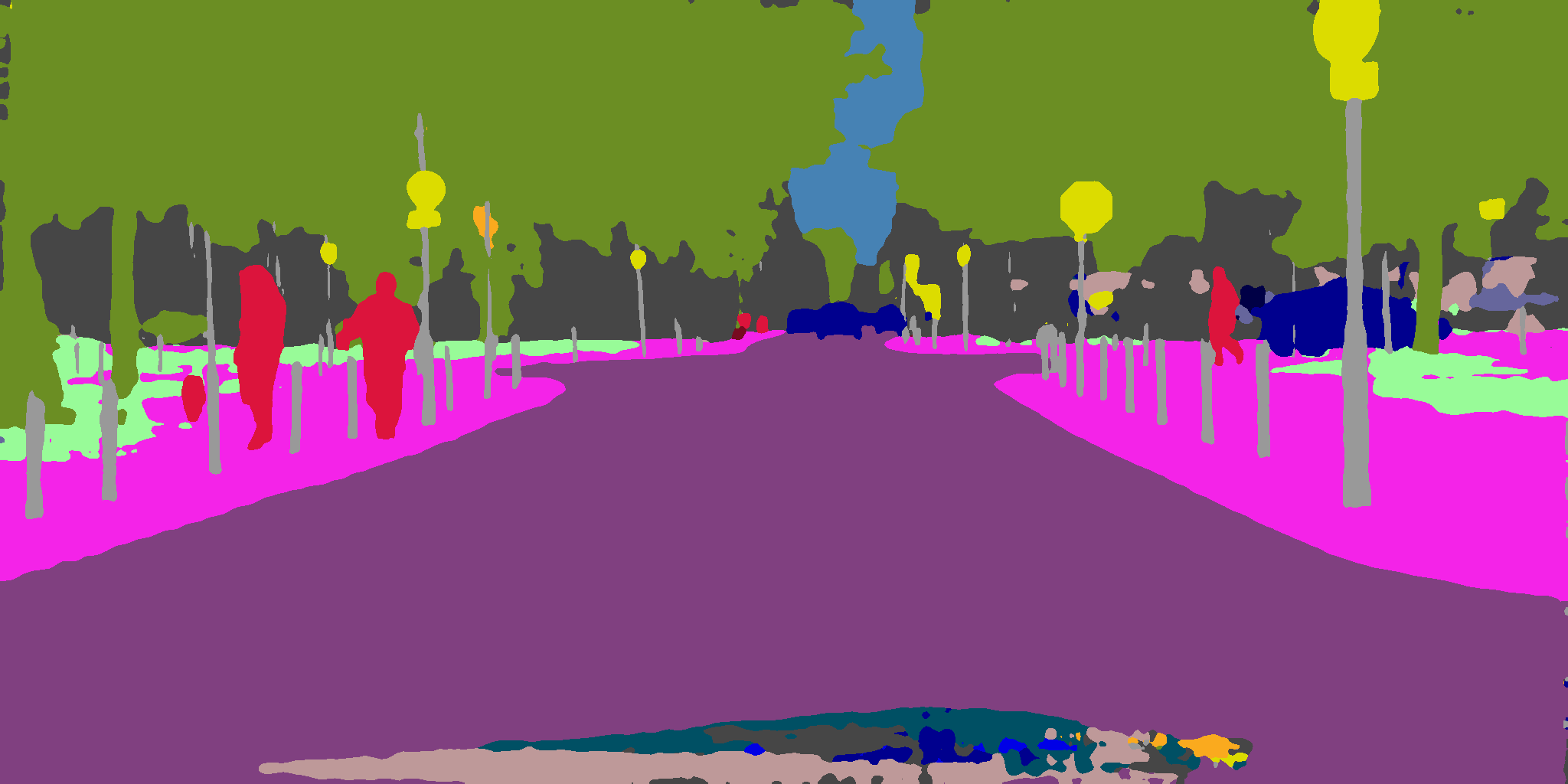}
            & \includegraphics[width=2.89em, height=1.9em]{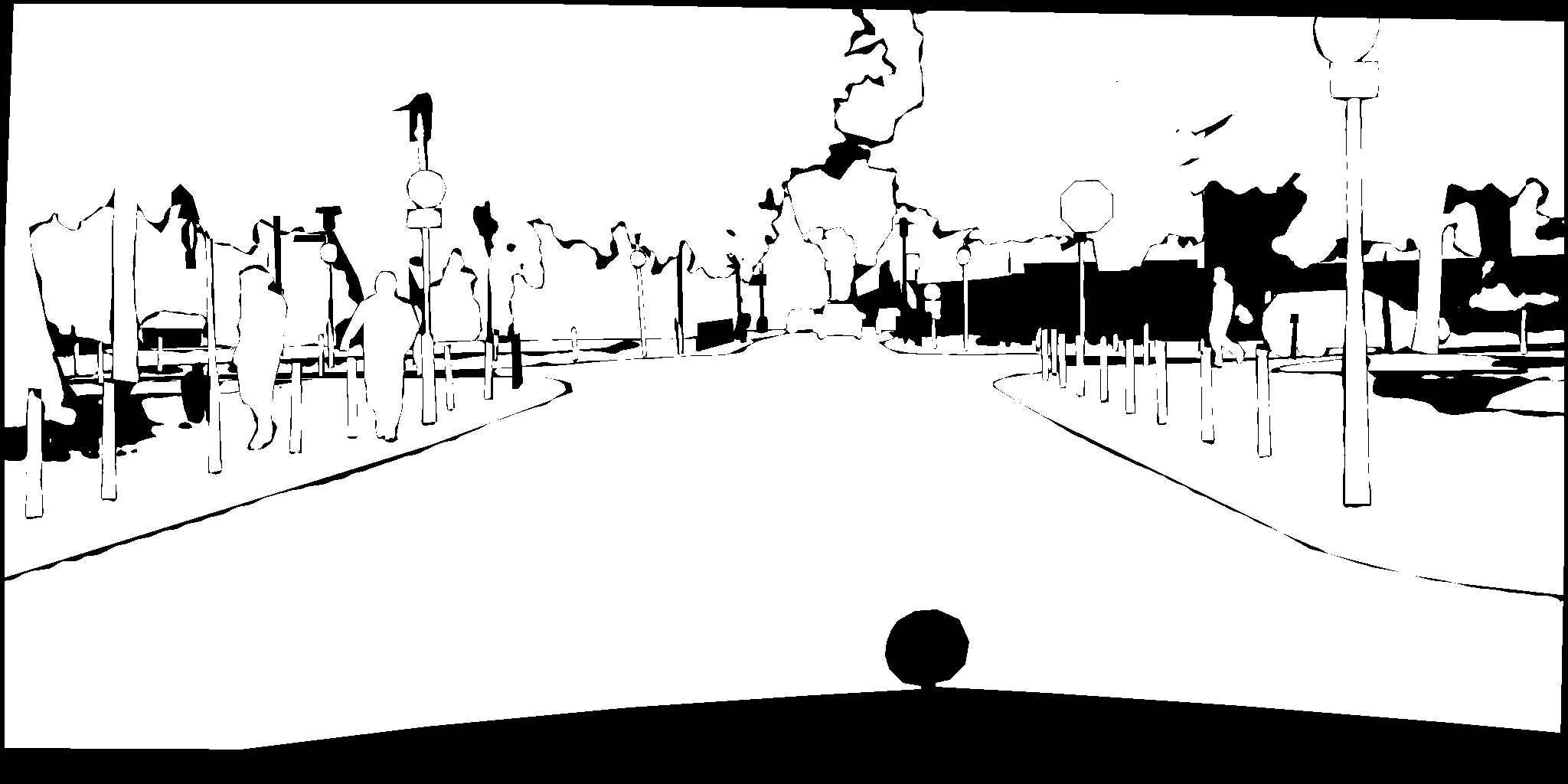} 
            & \includegraphics[width=2.89em, height=1.9em]{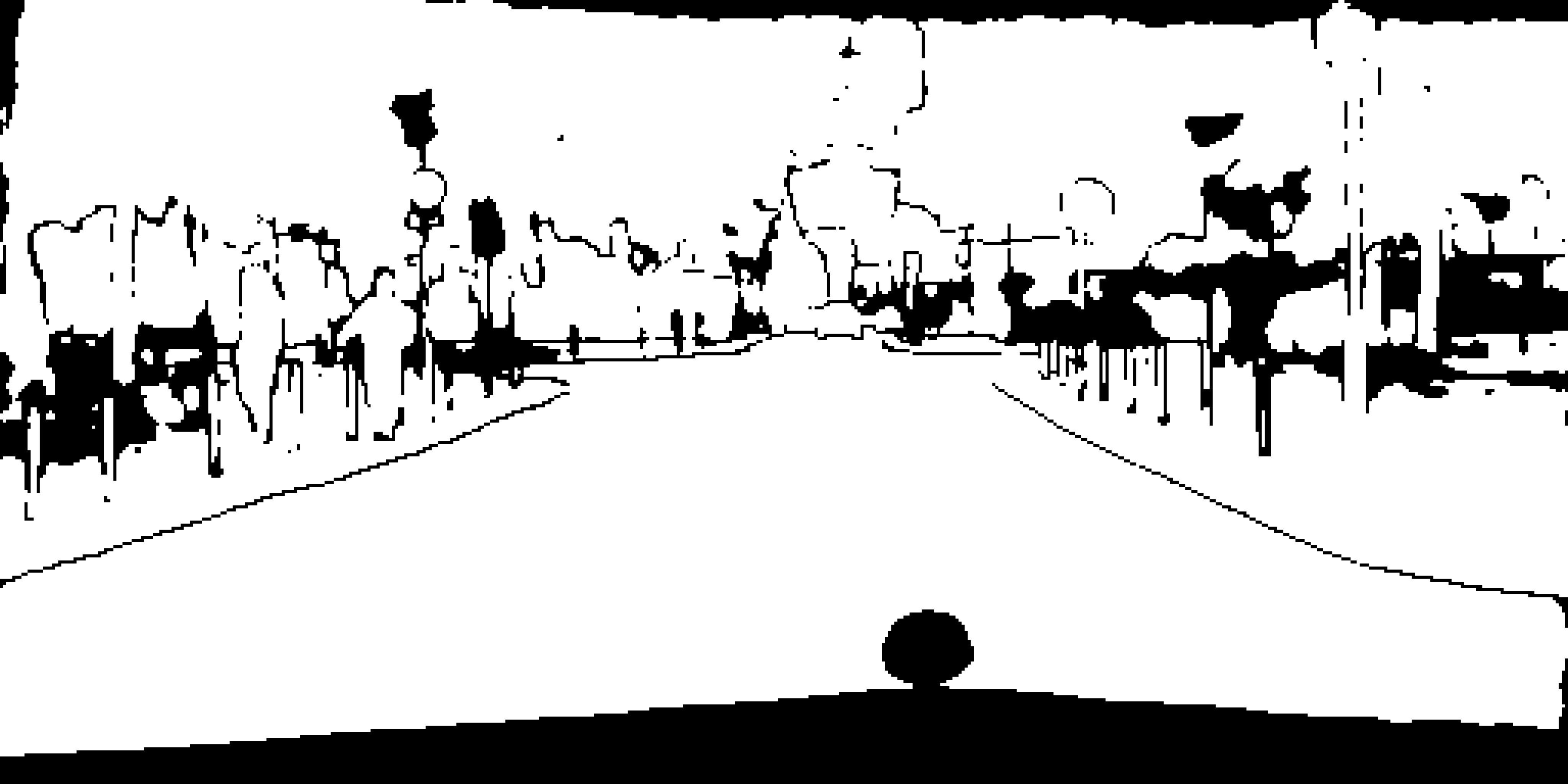}
            & \includegraphics[width=2.89em, height=1.9em]{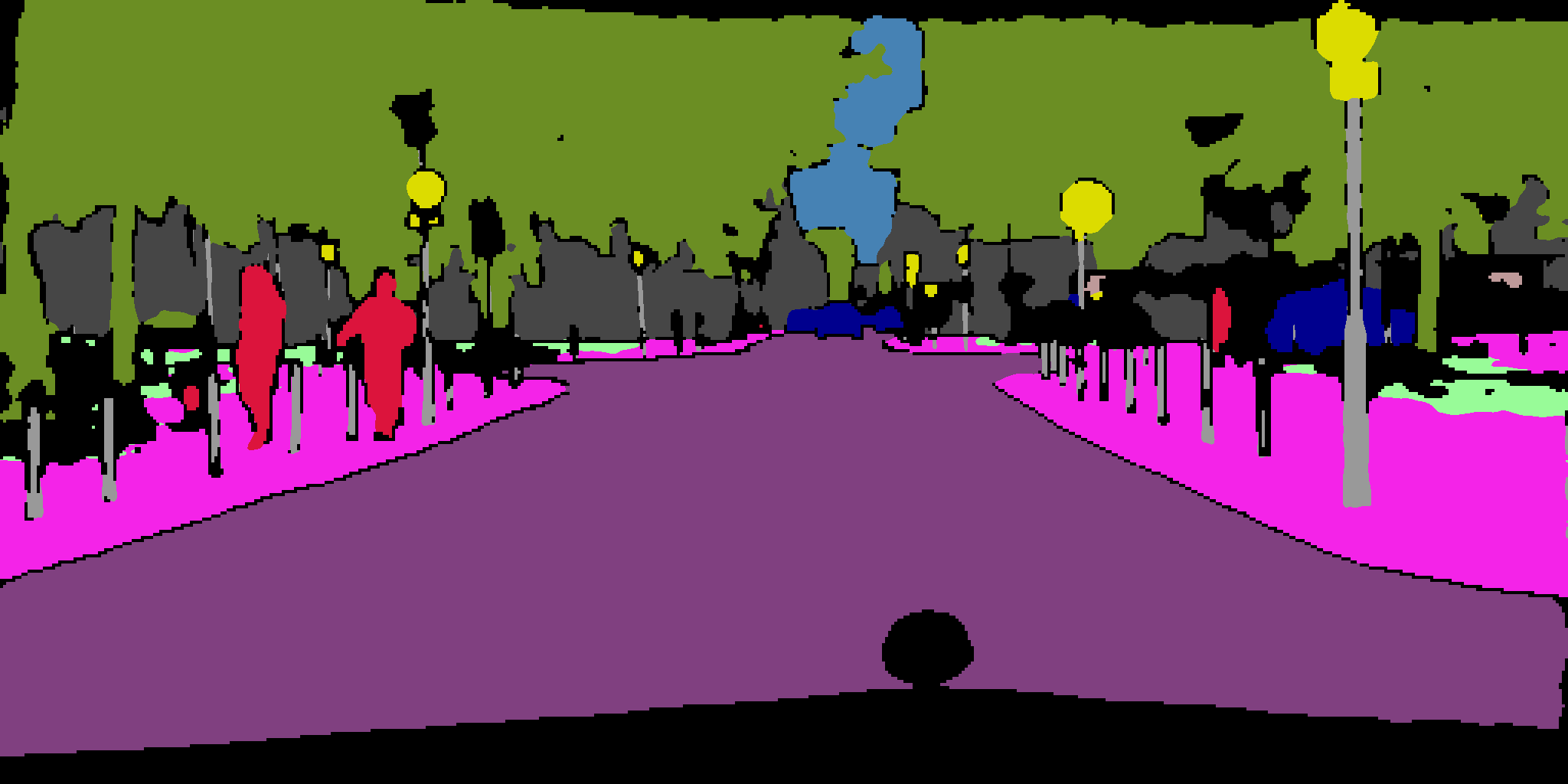}
            \\
            
            &&&&& \\
            &&&&& \\ 
            &&&&& \\ 
            &&&&& \\ 
              {\fontsize{3.5}{4.5}\selectfont Input Image}
            & {\fontsize{3.5}{4.5}\selectfont GT}
            & {\fontsize{3.5}{4.5}\selectfont (a)}
            & {\fontsize{3.5}{4.5}\selectfont (b)}
            & {\fontsize{3.5}{4.5}\selectfont (c)}
            & {\fontsize{3.5}{4.5}\selectfont (d)}
              
        \end{tabular}
    }
    \caption{Qualitative results on unlabeled data of training set on Cityscapes in the labeled ratio of 1/8. (a) Segmentation prediction from the main segmentation network. (b) Ground truth binary mask. (c) Binary mask predicted by ELN. (d) Filtered segmentation prediction by the predicted binary mask. Erroneous predictions colored in white in (d) are not used as pseudo labels.}
    \label{sec:qual-1-city}
\end{figure*}

\noindent\textbf{Performance analysis on error localization network.}
We further conduct additional experiments to compare our method with two similar approaches to demonstrate the effectiveness of ELN. We conduct experiments on the ratio of 1/20 to PASCAL VOC 2012 with ResNet-50 as a backbone network. As the first thing to compare, we consider a simple error correction network (\emph{s}-ECN) which has a similar learning strategy as ELN; \emph{s}-ECN is trained with pixel-wise cross-entropy loss and yields a corrected segmentation prediction as an output, not a binary mask. We choose another method, performing confidence score threshold on the output of segmentation prediction after softmax layer, without an additional network. As we can see from Table \hyperref[sec:ablation-1]{3}, ELN achieves the highest mIoU value over the other two approaches. We also conduct another experiment to understand how well each method performs error localization to unseen data. In Table \hyperref[sec:ablation-1-2]{4}, ELN shows the highest F1 score among methods. Note that results of \emph{s}-ECN are worse than Threshold; it emphasizes the limitations of the error correction scheme, implying that it does not work as intended due to its harsh training condition. In Fig. \hyperref[sec:qual-1-voc]{5} and \hyperref[sec:qual-1-city]{6}, we display our qualitative results of segmentation prediction and its binary mask.

\begin{table}[!t]
\centering

\begin{tabular}{@{}lcccl@{}}
\toprule
Method        & ELN                  & \emph{s}-ECN         & \multicolumn{1}{c}{Threshold}\\
\midrule
mIoU          & \textbf{70.52}       & 67.14       & \multicolumn{1}{c}{67.77}  \\
\bottomrule
\end{tabular}
\label{sec:ablation-1}
\caption{
mIoU value of ELN, \emph{s}-ECN and confidence score threshold method. The experiment is conducted in a \emph{val} set of PASCAL VOC 2012.
} 
\end{table}
\begin{table}[!t]
\centering

\begin{tabular}{@{}lcccl@{}}
\toprule
Method      & ELN          & \emph{s}-ECN         & \multicolumn{1}{c}{Threshold}\\
\midrule
Precision   & 0.6961       & 0.7060       & \multicolumn{1}{c}{0.7054}  \\
\midrule
Recall      & 0.9673       & 0.8294       & \multicolumn{1}{c}{0.8783}  \\
\midrule
F1 score    & \textbf{0.7881}       & 0.7424       & \multicolumn{1}{c}{0.7627}  \\
\bottomrule
\end{tabular}
\label{sec:ablation-1-2}
\caption{
Precision, Recall, and F1 score of the ELN, \emph{s}-ECN, and confidence score threshold method. Reported scores are averages of all the results of each image. To compare \emph{s}-ECN, we only consider its error localization ability, not a correction. The experiment is conducted on the unlabeled data of the given ratio of 1/20 to PASCAL VOC 2012.
} 
\end{table}



\begin{table}[!t]
\centering

\begin{tabular}{@{}lcccc@{}}
\toprule
\# of aux. decoders     & 0        & 1        & 2       & 3                \\
\midrule
$\alpha$    & -        & 20       & 20,50   & 20,50,100        \\
\midrule
mIoU                    & 69.89    & 70.20    & 70.52   & \textbf{71.13}  \\
\bottomrule
\end{tabular}
\label{sec:ablation-2}
\caption{
 An ablation study for the model performance according to the number of decoders and loss constrain parameter $\alpha$.
} 

\end{table}
\subsection{Ablation Studies}
We conduct ablation studies to investigate the impacts of each component of the proposed method. The experiment is based on PASCAL VOC 2012 with the ratio of 1/20, and results are averaged over three times. We use ResNet-50 as the backbone of the main segmentation network.

\noindent\textbf{Different number of auxiliary decoders.} The auxiliary decoder plays a critical role in ELN learning. 
We experiment with how the auxiliary decoders affect the overall performance by changing the number of auxiliary decoders and constraints parameters. The result is listed in Table \hyperref[sec:ablation-2]{5}. As reported from the experiments, the performance improves as the number of decoders increases and a high loss constrain value is applied. It shows that various quality of segmentation predictions helps the effective learning of the ELN. Note that we could achieve sufficient performance improvement with only two auxiliary decoders.

\begin{table}[!t]
\centering
\begin{tabular}{@{}cccc@{}}
\toprule
Threshold & $\mathcal{L}_{pseudo}$ & $\mathcal{L}_{contra}$ & $\mathcal{L}_{pseudo}+ \mathcal{L}_{contra}$  \\
\midrule
67.77 & 69.14 & 69.30 & \textbf{70.52}\\ 
\bottomrule
\end{tabular}
\label{sec:ablation-4}
\caption{
An ablation study on different loss combinations in mIoU.
``Threshold" in the table is the method using none of the two losses associated with ELN but applying a confidence score threshold.
} 
\end{table}

\noindent\textbf{Different loss combination in Eq.~\eqref{eqn:11}.} 
In the semi-supervised learning stage, $\mathcal{L}_{pseudo}$ and $\mathcal{L}_{contra}$ are jointly optimized. The proposed pixel-wise contrastive loss, $\mathcal{L}_{contra}$, enables the training of feature embeddings in more diverse contexts by learning the relation between images in a batch with a standard pixel-wise cross-entropy loss $\mathcal{L}_{pseudo}$. We make a comparison to investigate effect of each loss term in Eq.~\eqref{eqn:11}. As shown in Table \hyperref[sec:ablation-4]{6}, each term contributes to the performance, and using both of them improves the most.

\section{Conclusion}\label{sec:Conclusion}
We have presented a novel training framework suitable for semi-supervised semantic segmentation tasks. 
To mitigate the performance degradation caused by confirmation bias due to invalid pseudo labels, we have proposed error localization network (ELN) and its training scheme. 
Our experiments validated that ELN effectively removes error of pseudo labels for unseen data, which demonstrate that our learning strategy using erroneous predictions simulated by auxiliary decoders is helpful.
Our method achieved the state of the art on both of the PASCAL VOC 2012 and Cityscapes datasets with high generalization capability.

\vspace{2.7mm}
\noindent\textbf{limitations.} Due to the additional auxiliary networks, our method needs a relatively larger amount of GPU memory during training, and as the number of auxiliary decoders increases, larger memory footprint is required. ELN sometimes failed to indicate erroneous predictions that the main segmentation network has strong confidence (\ie, low-entropy). 

\vspace{2mm}
\noindent\textbf{Acknowledgement.}  
This work was supported by Samsung Electronics Co., Ltd (IO201210-07948-01). %

{
    
    \bibliographystyle{ieee_fullname}
    \bibliography{cvlab_kwak}
}

\end{document}